\documentclass[conference]{IEEEtran}
\IEEEoverridecommandlockouts
\usepackage[colorlinks=true,linkcolor=blue,anchorcolor=blue,citecolor=blue,filecolor=blue,menucolor=blue,runcolor=blue,urlcolor=blue,breaklinks]{hyperref}
\usepackage{cite}
\usepackage{amsmath,amssymb,amsfonts}
\usepackage{algorithmic}
\usepackage{graphicx}
\usepackage{textcomp}
\usepackage{comment}
\usepackage{xcolor}
\def\BibTeX{{\rm B\kern-.05em{\sc i\kern-.025em b}\kern-.08em
    T\kern-.1667em\lower.7ex\hbox{E}\kern-.125emX}}

\usepackage{tikz}
\newcommand{\greenline}{\raisebox{2pt}{\tikz{\draw[-,black!50!green,solid,line width = 1pt](0,0) -- (3mm,0);}}}
\newcommand{\redline}{\raisebox{2pt}{\tikz{\draw[-,red!100!orange,solid,line width = 1pt](0,0) -- (3mm,0);}}}
\newcommand{\orangeline}{\raisebox{2pt}{\tikz{\draw[-,yellow!30!orange,solid,line width = 0.9pt](0,0) -- (3mm,0);}}}

\begin{document}

%Robust Data Management and Storage Based on Optimal Sensor Placement in Big Data

%Robust Data Management and Storage Using Optimal Sensor Placement in Big Data

%Computationally and memory-efficient robust predictive analytics using big data

%Efficient Big Data Predictive Analytics: Optimizing Computation and Memory

%Optimizing Big Data Workflows: Techniques in Preprocessing, Compression, and Predictive Modeling

\title{Computationally and Memory-Efficient Robust Predictive Analytics Using Big Data\\
%{\footnotesize \textsuperscript{*}Note: Sub-titles are not captured in Xplore and should not be used}

\thanks{
This work is part of SFI AutoShip, an 8-year research-based innovation center. 
In addition, this research project is integrated into the PERSEUS doctoral program. 
We want to thank our partners, including the Research Council of Norway, under project number 309230, and the European Union’s Horizon 2020 research and innovation program under the Marie Skłodowska-Curie grant agreement number 101034240. Furthermore, we thank Idletechs AS for providing us with the thermal camera data.
\\
\\
Copyright © 2024 IEEE. This article has been accepted for publication in a future issue of the 2024 IEEE Conference on Artificial Intelligence.}
}
%\author{\IEEEauthorblockN{Anonymous Authors}}
%\begin{comment}
\author{\IEEEauthorblockN{1\textsuperscript{st} Daniel Menges}
\IEEEauthorblockA{\textit{Department of Engineering Cybernetics} \\
\textit{Norwegian University of Science and Technology}\\
Trondheim, Norway \\
daniel.menges@ntnu.no}
\and
\IEEEauthorblockN{2\textsuperscript{nd} Adil Rasheed}
\IEEEauthorblockA{\textit{Department of Engineering Cybernetics} \\
\textit{Norwegian University of Science and Technology}\\
Trondheim, Norway \\
adil.rasheed@ntnu.no}
}
%\end{comment}

\maketitle

\begin{abstract}
In the current data-intensive era, big data has become a significant asset for Artificial Intelligence (AI), serving as a foundation for developing data-driven models and providing insight into various unknown fields. This study navigates through the challenges of data uncertainties, storage limitations, and predictive data-driven modeling using big data. We utilize Robust Principal Component Analysis (RPCA) for effective noise reduction and outlier elimination, and Optimal Sensor Placement (OSP) for efficient data compression and storage. The proposed OSP technique enables data compression without substantial information loss while simultaneously reducing storage needs.
While RPCA offers an enhanced alternative to traditional Principal Component Analysis (PCA) for high-dimensional data management, the scope of this work extends its utilization, focusing on robust, data-driven modeling applicable to huge data sets in real-time. 
For that purpose, Long Short-Term Memory (LSTM) networks, a type of recurrent neural network, are applied to model and predict data based on a low-dimensional subset obtained from OSP, leading to a crucial acceleration of the training phase. LSTMs are feasible for capturing long-term dependencies in time series data, making them particularly suited for predicting the future states of physical systems on historical data.
All the presented algorithms are not only theorized but also simulated and validated using real thermal imaging data mapping a ship's engine.
\end{abstract}


\section{Introduction}
In the context of Artificial Intelligence (AI), data have taken center stage, influencing decision-making processes in many domains, from healthcare \cite{raghupathi_big_2014} to econometrics \cite{varian_big_2014}, manufacturing \cite{nagorny_big_2017}, and more. However, while big data offers incredible potential, it is essential to understand its strengths and inherent flaws, especially since data can be erroneous due to various factors such as sensor uncertainties and transmission errors.
Therefore, data can sometimes be misinterpreted if not used appropriately, particularly when the underlying data are flawed or inaccurate \cite{pitici_rise_2014}.
The ability to effectively handle, analyze, and interpret these growing volumes of data is essential. Therefore, the development and deployment of robust data analysis techniques is of critical importance. %Surveys assessing various preprocessing methods \cite{fan_review_2021} mainly cover statistical methods, such as Generalized Extreme Studentised Deviate (GESD), order-based, and clustering-based methods, notably the DBSCAN algorithm. Their suitability varies based on data characteristics and the application's specific requirements. 

Among various available data analysis tools, Principal Component Analysis (PCA) \cite{jolliffe_principal_2002} has gained significant attention due to its ability to reduce the dimensionality of data sets while retaining most of the underlying information \cite{abdi_principal_2010}. However, traditional PCA is highly susceptible to outliers and corruptions in the data, which can substantially impact its performance and the accuracy of subsequent analyses. Consequently, there is a need for more robust techniques that can handle such irregularities. Robust Principal Component Analysis (RPCA), an advanced variant of PCA, offers more reliable results by robustly separating low-rank and sparse components in the data, even in the presence of outliers and corruptions \cite{hubert_robpca_2005}. 
The concept of RPCA for decomposing a data matrix in a low-rank and a sparse component is accurately described in \cite{candes_robust_2011}. The decomposed components use a convex program called Principal Component Pursuit. The method, which can recover the principal components even when data entries are corrupted or missing, has applications in video surveillance for object detection, in cluttered backgrounds and face recognition for removing shadows, in specularities, and more. A detailed comparison of PCA and RPCA is given in \cite{scherl_robust_2019}, showcasing the benefits and robustness of RPCA.

In parallel, considering the growing need for big data, one of the key challenges that emerges is the efficient storage and transmission of these enormous volumes of data. A novel approach to this problem is the concept of Optimal Sensor Placement (OSP) \cite{manohar_data-driven_2018}. OSP involves strategic positioning of sensors to capture the most relevant data, significantly reducing redundancy and facilitating efficient data storage and transmission. In essence, OSP aims to obtain a compressed version of the data without a significant loss of information.

%In addition to this, accurately estimating and mitigating the effects of noise in collected data is crucial to achieve precise and reliable results. Noise can distort the real patterns in data and lead to misleading results. Noise estimation techniques, therefore, play a vital role in preprocessing data, improving the quality of data by identifying and reducing the impact of noise.

%In our pursuit of creating sophisticated data-driven models, we harness the power of Long Short-Term Memory (LSTM) networks. Specifically, we focus on the few pivotal pixel measurements procured through the Optimal Sensor Placement (OSP). Leveraging LSTM on this compressed representation not only streamlines the model but also accelerates the training phase significantly. This expedited training process makes the approach especially relevant and versatile, catering to a myriad of applications.

%Once the predictions are made on these select pixel measurements, a crucial step ensues: reconstructing the entire images. Utilizing the predicted pixel values, we meticulously reconstruct the thermal images, thereby enabling us to make informed predictions about the thermal states of the ship's engine. This synergy between OSP and LSTM, combined with the subsequent image reconstruction, embodies a holistic approach to forecasting, ensuring both efficiency and accuracy in predicting the engine's thermal dynamics.

Through a comprehensive examination of RPCA and OSP, this study aims to explore the synergies among these methodologies and their collective impact on improving the accuracy and efficiency of big data modeling and analysis.

%Through a comprehensive examination of RPCA and OSP, this study embarks on a deep exploration of the combined strengths of these methodologies. Our primary goal is to unearth the synergies among them and gauge their collective efficacy in enhancing the precision and efficiency of big data modeling and analysis.

Furthermore, we extend this work by integrating a data-driven modeling approach for real-time predictions using Long Short-Term Memory (LSTM) networks, which was first proposed by \cite{hochreiter_long_1997}. The specialized design of LSTMs, with its gate mechanisms, allows them to learn long-term dependencies in the data \cite{chung_gated_2015}. Artificial Neural Networks (ANNs) have gained considerable traction in various forecasting domains due to their adaptability, nonlinearity, and the capability to map arbitrary functions. However, they require a lot of computational time for training \cite{zhang_forecasting_1998}.
Therefore, we create LSTM models based on the few selected data points obtained from the OSP algorithm. This technique significantly accelerates the training phase, making the proposed methodology adaptable to a wide range of applications.
Once these few data points (measurements) are predicted using LSTMs, we reconstruct the full data dimension via the concept of OSP, allowing predictions of future states in full dimension with remarkable accuracy. The integration of RPCA, OSP, and LSTM offers a novel approach to big data modeling, promising both robustness and scalability in various real-world scenarios.

In this study, we applied the algorithms on a dataset from a thermal camera mapping a ship's engine. The thermal images provided insight into the temperature profiles and fluctuations, offering a unique perspective on the engine's operational behavior and performance. 
Condition monitoring is crucial to maintain safe maritime operations \cite{mohanty_machinery_2014} and can provide insight into the reliability of a vessel's engine and its components. By identifying anomalies early on, it is possible to predict the lifespan of these components and prevent significant breakdowns. As pointed out in \cite{inproceedings}, the maritime sector rarely employs predictive maintenance. Instead, most maintenance activities on ships tend to be preventive. This frequently leads to higher costs as replaced components might have had a longer usable life endurance.

In summary, this study addresses three core challenges:
\begin{itemize}
    \item The robust treatment of uncertainties such as outliers and corruptions in data due to the use of affordable, nonintrusive thermal camera measurements.
    \item The requirement for memory-efficient storage techniques due to the vast data generated.
    \item The capability of proactive maintenance in real-time through predictive data-driven modeling.
\end{itemize}

%Finally, we incorporate Dynamic Mode Decomposition (DMD) which offers the capability to build predictive models based on the temporal evolution of modes, which is powerful for analyzing historical data. However, its real-time applicability for big data processing can be challenging due to the high computational requirements. To circumvent this issue, our study applies DMD to a lower-dimensional space that is carefully extracted using the OSP technique. This approach is developed and precisely explained in \cite{menges_arxiv}.
%This combined application of OSP and DMD not only optimizes the processing of data, making it feasible for real-time analysis but also enhances the prediction performance. The reason is that OSP, guided by an information-theoretic approach, ensures that the extracted lower-dimensional space retains the most significant features and patterns of the original data. Therefore, the predictive model built by DMD on this reduced dataset still encompasses the main dynamics of the full-scale system.
%This innovative approach offers a substantial improvement in the real-time capabilities of data modeling and prediction while maintaining high prediction accuracy. This is particularly crucial for applications such as thermal camera data analysis of a ship's engine, where real-time monitoring and predictive maintenance can lead to more efficient and safer operations.

\section{Theory}
This section provides a detailed overview of the statistical techniques used in this study. We introduce the concept of Principal Component Analysis (PCA) and its robust counterpart, Robust Principal Component Analysis (RPCA), for data cleaning. Furthermore, the section covers the idea of Optimal Sensor Placement (OSP) used for effective data compression and storage management. 
%Lastly, we discuss the implementation of Dynamic Mode Decomposition (DMD) within a lower-dimensional space rendered by OSP, a strategy devised to enable real-time prediction performance.
\subsection{Principal Component Analysis}
%Principal Component Analysis (PCA) is a statistical method that uses an orthogonal transformation to convert a set of observations of possibly correlated variables into a set of values of linearly uncorrelated variables called principal components.
%In a real space with $p$ dimensions, a group of points has $p$ principal components. These components are a sequence of direction vectors. Each vector, starting from the $i^{th}$ one, indicates the direction of a line that is the best fit for the data, and this line is also orthogonal, or perpendicular, to the preceding principal components. Here, a best-fitting line is defined as one that minimizes the average squared distance from the points to the line.
Principal Component Analysis (PCA) is a statistical procedure that uses an orthogonal transformation to convert a set of observations of possibly correlated variables into a set of linearly uncorrelated variables, termed principal components. This procedure allows for identifying the directions (principal components) where the data vary the most. There are two main approaches to compute the PCA. The eigenvector approach and the Singular Value Decomposition (SVD) approach.
The general concepts are described in detail in \cite{shlens_tutorial_2014}. The SVD approach is often chosen since it is numerically more robust.
\\

\subsubsection*{Singular Value Decomposition Approach}
PCA is closely related to SVD, a factorization of a real or complex matrix. For any real matrix $\mathbf{A}\in \mathbb{R}^{m\times n}$, with $m \geq n$, there exists a factorization of the form 
\begin{equation}
\mathbf{A} = \mathbf{U} \mathbf{\Sigma} \mathbf{V}^T,
\end{equation}
where $\mathbf{U}\in \mathbb{R}^{m\times m}$, $\mathbf{\Sigma}\in \mathbb{R}^{m\times n}$, and $\mathbf{V}\in \mathbb{R}^{n\times n}$.
The columns of $\mathbf{U}$ are orthonormal eigenvectors of $\mathbf{AA}^T$, and the columns of $\mathbf{V}$ are orthonormal eigenvectors of $\mathbf{A}^T\mathbf{A}$. The diagonal elements of $\mathbf{\Sigma}$ are the square roots of the eigenvalues of $\mathbf{A}^T\mathbf{A}$ (or equivalently, $\mathbf{AA}^T$), and are called the singular values of $\mathbf{A}$.
To see this, we first consider the matrix $\mathbf{A}^T\mathbf{A}$, which is a symmetric matrix. By the spectral theorem, we can factorize it as
\begin{equation}
\mathbf{A}^T\mathbf{A} = \mathbf{V} \mathbf{\Sigma}^2\mathbf{V}^T.
\end{equation}
Similarly, we can factorize $\mathbf{AA}^T$ as
\begin{equation}
\mathbf{AA}^T = \mathbf{U} \mathbf{\Sigma}^2 \mathbf{U}^T.
\end{equation}
Using these two identities, it can be shown that
\begin{equation}
\mathbf{A} = \mathbf{U} \mathbf{\Sigma} \mathbf{V}^T,
\end{equation}
which is the Singular Value Decomposition of $\mathbf{A}$.

Consider a data matrix $\mathbf{X} \in \mathbb{R}^{m \times n}$, where each row is an observation and each column is a variable. We assume that the data have been centered, i.e. the column means have been subtracted off.

\begin{enumerate}
\item \textit{Perform a lower-ranked SVD:} Compute the SVD of $\mathbf{X}$ by $\mathbf{X} = \mathbf{U}_r\mathbf{\Sigma}_r\mathbf{V}_r^T+\mathbf{E}$. Here, $\mathbf{U}_r \in \mathbb{R}^{m \times r}$ and $\mathbf{V}_r^\top \in \mathbb{R}^{r \times n}$ are orthogonal matrices containing left and right singular vectors and $r$ is the number of principal components, respectively. The matrix $\mathbf{\Sigma}_r \in \mathbb{R}^{r \times r}$ contains the $r$ largest singular values in decreasing order on the diagonal. Furthermore, the matrix $\mathbf{E}$ contains the unmodeled residuals due to the dimensionality reduction.

\item \textit{Principal components:} Finally, the principal components of $\mathbf{X}$ are given by $\mathbf{X}\mathbf{V}_r \approx \mathbf{U}_r \mathbf{\Sigma}_r$. The $i$-th column of $\mathbf{X}\mathbf{V}_r$ is the projection of the data onto the $i$-th principal direction (i.e., the $i$-th eigenvector).
\end{enumerate}

This process shows how PCA can be derived from the SVD of a data matrix. However, traditional PCA is highly sensitive to outliers and data corruptions.

\subsection{Robust Principal Component Analysis} \label{sec:RPCA}
The most significant advantage of RPCA over standard PCA is its resilience to outliers. Traditional PCA is sensitive to outliers because it tries to find a lower-dimensional representation that best explains the variance in the data. If outliers are present, PCA may be heavily influenced by them, leading to a representation that does not accurately capture most of the underlying structure of the data. RPCA, on the other hand, explicitly models these outliers, resulting in a more accurate and robust representation of the primary data structure.

In certain contexts, RPCA can better recover the true underlying low-rank structure of data compared to PCA, especially when the data are grossly corrupted or when a significant amount of data is missing.

RPCA works by decomposing the data matrix into a low-rank matrix and a sparse matrix. The low-rank matrix captures the principal components, and the sparse matrix captures outliers or anomalies. This separation can be very useful in many applications, such as image and video processing, where the low-rank component can correspond to the background and the sparse component can correspond to moving objects.
The general idea is to decompose the data matrix $\mathbf{X}$ into two components expressed by
\begin{equation}
    \mathbf{X} = \mathbf{L} + \mathbf{S}.
\end{equation}
Here, the matrix $\mathbf{L}$ describes the low-rank matrix that captures the main structure of the data, while the matrix $\mathbf{S}$ is sparse and captures outliers and corruptions. Therefore, the goal is to find $\mathbf{L}$ and $\mathbf{S}$ which satisfy 
\begin{equation}
\begin{split}
& \underset{\mathbf{L}, \mathbf{S}}{\text{minimize}} \hspace{0.5cm}\mathrm{rank}(\mathbf{L}) + \|\mathbf{S}\|_0, \\
& \text{subject to} \hspace{0.5cm} \mathbf{L} + \mathbf{S} = \mathbf{X},
\end{split}
\label{eq:RPCA_ideal}
\end{equation}
where $\|\mathbf{S}\|_0$ describes the zero norm of $\mathbf{S}$, and $\mathrm{rank}(\mathbf{L})$ specifies the rank of $\mathbf{L}$. However, due to the nonconvex nature of both the rank($\mathbf{L}$) and the $\|\mathbf{S}\|_0$, this optimization problem becomes intractable \cite{scherl_robust_2019}. To overcome this issue, convex relaxation \cite{JMLR:v11:zhang10a} provides an approach to approximate convexity for nonconvex problems. Convex relaxation allows transforming \eqref{eq:RPCA_ideal} into
\begin{equation}
\begin{split}
& \underset{\mathbf{L}, \mathbf{S}}{\text{minimize}} \hspace{0.5cm}\|\mathbf{L}\|_* + \lambda \|\mathbf{S}\|_1, \\
& \text{subject to} \hspace{0.5cm} \mathbf{L} + \mathbf{S} = \mathbf{X},
\end{split}
\label{eq:PCP}
\end{equation}
where $\|\cdot\|_1$ is the $L_1$ norm given by the sum of the absolute values of the matrix entries, $\|\cdot\|_*$ is the nuclear norm given by the sum of singular values, and $\lambda$ is a hyperparameter. While minimization of $\|\mathbf{S}\|_1$ leads to an approximation of minimizing $\|\mathbf{S}\|_0$, minimization of $\|\mathbf{L}\|_*$ leads to an approximation of the lowest possible $\mathrm{rank}(\mathbf{L})$. The problem described in \eqref{eq:PCP} is convex and known as Principal Component Pursuit (PCP). To solve this convex problem, the Augmented Lagrange Multiplier (ALM) algorithm is suggested \cite{lin_augmented_2010}. The augmented lagrange multiplier can be formulated as
\begin{equation}
    \hspace{-0.7em}\resizebox{.93\hsize}{0.012\vsize}{$\mathcal{L}(\mathbf{L},\mathbf{S},\mathbf{\Lambda})=\|\mathbf{L}\|_* + \lambda \|\mathbf{S}\|_1+\langle \mathbf{\Lambda}, \mathbf{X} - \mathbf{L} - \mathbf{S} \rangle + \frac{\mu}{2}\|\mathbf{X}-\mathbf{L}-\mathbf{S}\|_{F}^2$}, \label{eq:ALM}
\end{equation}
where $\mathbf{\Lambda}$ is the matrix of Lagrange multipliers, $\mu$ is a hyperparameter, $\langle \cdot \rangle$ denotes the inner product, and $\|\cdot\|_F$ is the Frobenius norm, also known as the Euclidean norm, which is a measure of the magnitude or length of a matrix. Subsequently, we minimize $\mathcal{L}$ to solve for $\mathbf{L}_k$ and $\mathbf{S}_k$ at timestep $k$, where the matrix of Lagrange multipliers is updated by
\begin{equation}
    \mathbf{\Lambda}_{k+1} = \mathbf{\Lambda}_{k} + \mu(\mathbf{X}-\mathbf{L}_k-\mathbf{S}_k).
\end{equation}
As a result, RPCA decomposes a data matrix $\mathbf{X}$ into a low-rank component $\mathbf{L}$ and a sparse component $\mathbf{S}$.

\subsection{Optimal Sensor Placement} \label{sec:OSP}
Optimal Sensor Placement (OSP) is a method to identify the most insightful locations within a system for the positioning of sensors. This approach can maximize the measurements' entropy while minimizing the amount of sensors required. Here, entropy describes the abundance of information within a system.

Let $\boldsymbol{x} \in \mathbb{R}^n$ be a single data point in time, which can be approximated as
\begin{equation}
\boldsymbol{x} \approx \mathbf{\Psi}_r \boldsymbol{a},
\end{equation}
where $\boldsymbol{a} \in \mathbb{R}^{r}$ contain the coefficients that vary over time while the columns of $\mathbf{\Psi}_r$ are the modes of the lower-ranked Proper Orthogonal Decomposition (POD). POD is very similar to PCA. However, POD modes are not scaled by the singular value matrix $\mathbf{\Sigma}$, such as the principal components of PCA. Therefore, $\mathbf{\Psi}_r = \mathbf{U}_r$. If we assume that the measurements can be expressed by
\begin{equation}
\boldsymbol{y} = \mathbf{C}\boldsymbol{x},
\end{equation}
with $\mathbf{C}\in \mathbb{R}^{s\times n}$ being a sparse measurement matrix and $s$ the number of sensors, the measurements can be approximated by
\begin{equation}
\boldsymbol{y} \approx \mathbf{C}\mathbf{\Psi}_r \boldsymbol{a}.
\end{equation}
If we denote $\mathbf{\Theta} = \mathbf{C}\mathbf{\Psi}_r$, the estimated coefficients can be represented by
\begin{equation}
\boldsymbol{\hat{a}} = \mathbf{\Theta}^\dagger\boldsymbol{y}. \label{eq:a_est}
\end{equation}
Hence, we can derive an estimate of $\boldsymbol{x}$ yielding
\begin{equation}
\boldsymbol{\hat{x}} = \mathbf{\Psi}_r\boldsymbol{\hat{a}} = \mathbf{\Psi}_r(\mathbf{C}\mathbf{\Psi}_r)^\dagger\boldsymbol{y}. \label{eq:OSP_reconstruction}
\end{equation}
As $\mathbf{\Psi}_r$ can be determined using the lower-ranked SVD, the only unknown entity is the sparse measurement matrix $\mathbf{C}$. As described by \cite{manohar_data-driven_2018}, optimal sensor placement can be achieved by applying QR factorization with column pivoting to the POD modes $\mathbf{\Psi}_r$. In this relation, it is important to note that the number of sensors $s$ must satisfy $s\geq r$.

\section{Methodology}
This section describes a potential big data workflow for data cleaning, data compression, and computationally efficient data-driven modeling. The core of the proposed framework is built by RPCA, OSP, and LSTMs.

\subsection{Data Cleaning}
We use RPCA for data cleaning, introduced in Section~\ref{sec:RPCA}. The tuning parameters described in \eqref{eq:ALM} were chosen as $\lambda = 0.006$ and $\mu = 10^{-5}$.
After obtaining the decomposition of the data matrix $\mathbf{X}$ into $\mathbf{L}$ (low-rank matrix) and $\mathbf{S}$ (sparse matrix), a cleaned version of the data can be reconstructed. The low-rank matrix $\mathbf{L}$ represents the underlying physics, while the sparse matrix $\mathbf{S}$ contains anomalies and perturbations.
As a result, the matrix $\mathbf{L}$ represents a cleaned version of the data matrix $\mathbf{X}$.

\subsection{Data Compression}
To compress the data while simultaneously retaining the essential information about the underlying system, we apply OSP described in Section~\ref{sec:OSP} to the cleaned data matrix $\mathbf{L}$ obtained from RPCA.
The fundamental principle behind OSP is to maximize the fidelity of the data while minimizing the number of sensors or data points. By placing sensors in locations that capture the most variance or information in the data, we can represent the original high-dimensional data $\mathbf{X}$ with a significantly smaller set of measurements $\mathbf{Y}$, where $\mathbf{Y}$ contains $\boldsymbol{y}$ stacked over a specific historical window. This smaller set of measurements is represented by the sparse measurement matrix $\mathbf{C}$. The selected measurements or sensors produce a compressed version of the original data. By reducing the number of required sensors, OSP can lead to significant cost savings in scenarios where sensor deployment is expensive.

\subsection{LSTM-based Modeling of Sparse Measurements}
In the field of data-driven modeling, the power of neural networks, particularly LSTM networks, has been proven in many applications. LSTMs are designed to remember patterns over long sequences, making them suitable for modeling time-series data. However, LSTMs may not be computationally suitable for large datasets. Therefore, we apply LSTM to the lower-dimensional subset $\mathbf{Y}$ obtained from OSP.
The combination of LSTMs and OSP can drastically reduce the computational costs required to train LSTM networks.  When we use LSTMs to model these sparse measurements selected by OSP, we aim to capture the underlying temporal dynamics. Once trained, these networks can be used to predict the sparse data points. By subsequently applying the reconstruction algorithm given by \eqref{eq:OSP_reconstruction}, we can transform these sparse predictions into full-sized sensor space, mapping the original data dimensions. Note that if the data is sampled with an inconsistent frequency, an initial interpolation of the data can lead to more accurate models.

\subsection{Big Data Workflow}
\begin{figure}
    \centering
    \includegraphics[width=0.8\linewidth]{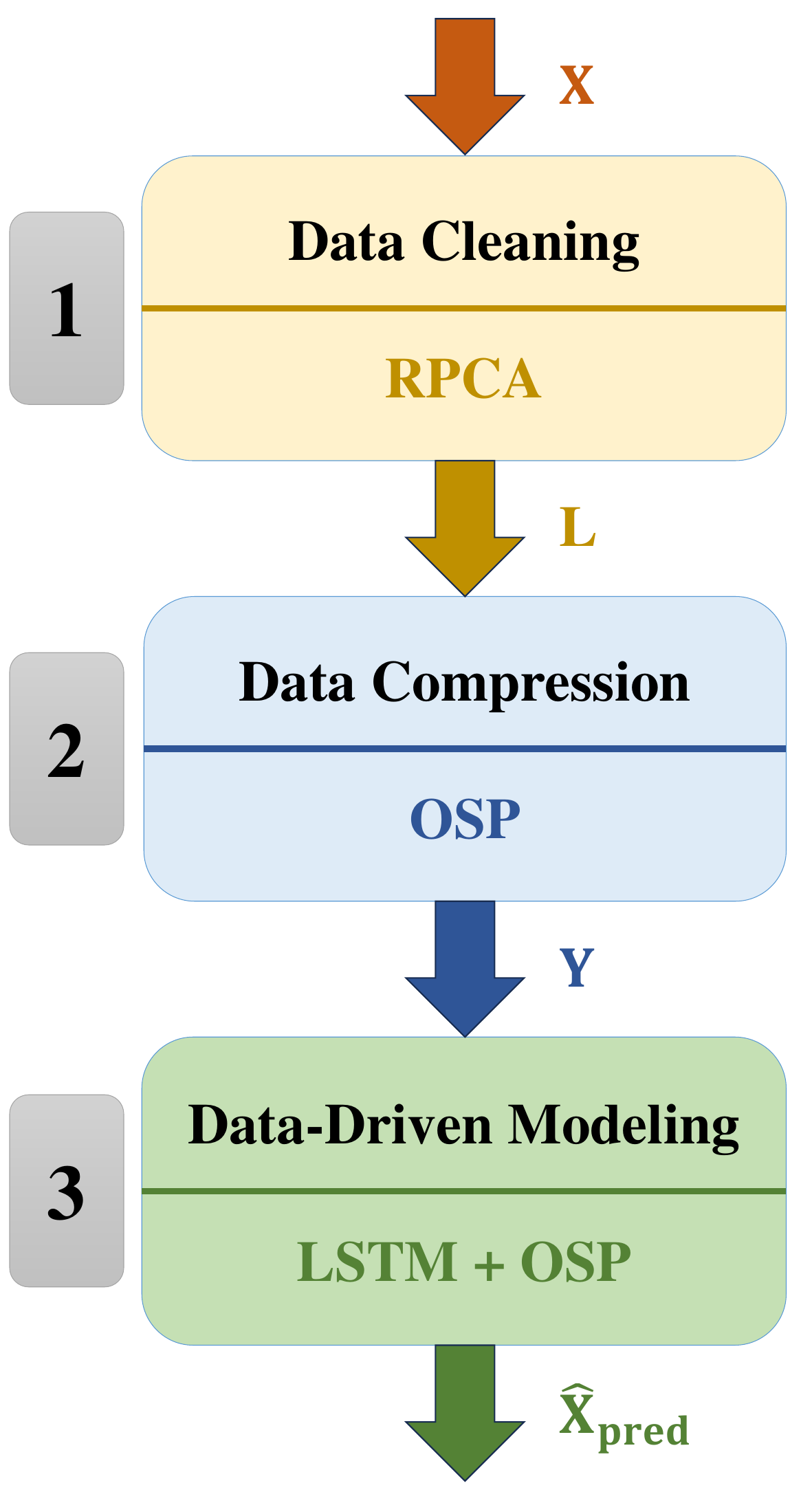}
    \caption{Concept of cleaning, compressing, and modeling the data.}
    \label{fig:framework}
\end{figure}

The previously described approaches can interact to combine their strengths into an optimized big data workflow. In Fig.~\ref{fig:framework}, we demonstrate a potential framework, employable to various applications for data preprocessing, compression, and modeling. The workflow has the following structure:
\begin{enumerate}
    \item \textbf{Data Cleaning:} RPCA generates a cleaned version $\mathbf{L}$ of the data matrix $\mathbf{X}$. Since $\mathbf{L}$ contains the information of interest (e.g., the underlying dynamics of the system), $\mathbf{L}$ can be propagated to subsequent processing and analysis methods.
    \item \textbf{Data Compression:} The OSP algorithm enables drastic compression of the cleaned data matrix $\mathbf{L}$. Computing the POD modes $\mathbf{\Psi}_r$ of $\mathbf{L}$ and finding the sparse measurement matrix $\mathbf{C}$, a small subset $\mathbf{Y}$ can be sufficient for data storage. The subset $\mathbf{Y}$ can be forwarded for continuative analysis and modeling. Note that $\mathbf{\Psi}_r$ and $\mathbf{C}$ must also be stored to extend the subset $\boldsymbol{y}$ to its original dimension $\boldsymbol{\hat{x}}$ (see \eqref{eq:OSP_reconstruction}).
    \item \textbf{Data-Driven Modeling:} In this step, data-driven models of the propagated subset $\mathbf{Y}$ are built using an LSTM-based neural network. The built data-driven models of the subset can thus be used to predict future states. After predicting the future subset, predictions of the original data dimension $\mathbf{\hat{X}_{pred}}$ can be computed using $\mathbf{\Psi}_r$ and $\mathbf{C}$ from the OSP algorithm.
\end{enumerate}

\section{Simulation Setup}
This study uses data acquired from a thermal camera mapping a ship's engine. The data were provided by Idletechs AS. Since the data were uncorrupted and free from outliers, we simulated synthetic perturbations affecting the data specified below. In addition, we describe the LSTM neural network setup that we chose for this study.

\subsection{Data}
The dataset under examination is derived from thermal camera imagery, which captures the engine of a ship, specifically that of a ferry. The primary intent behind acquiring these images was to observe the thermal behavior of the engine during various operational states, including taking off, steady-state driving, and docking. 

\begin{figure}[b]
    \centering
    \includegraphics[width=0.6\linewidth]{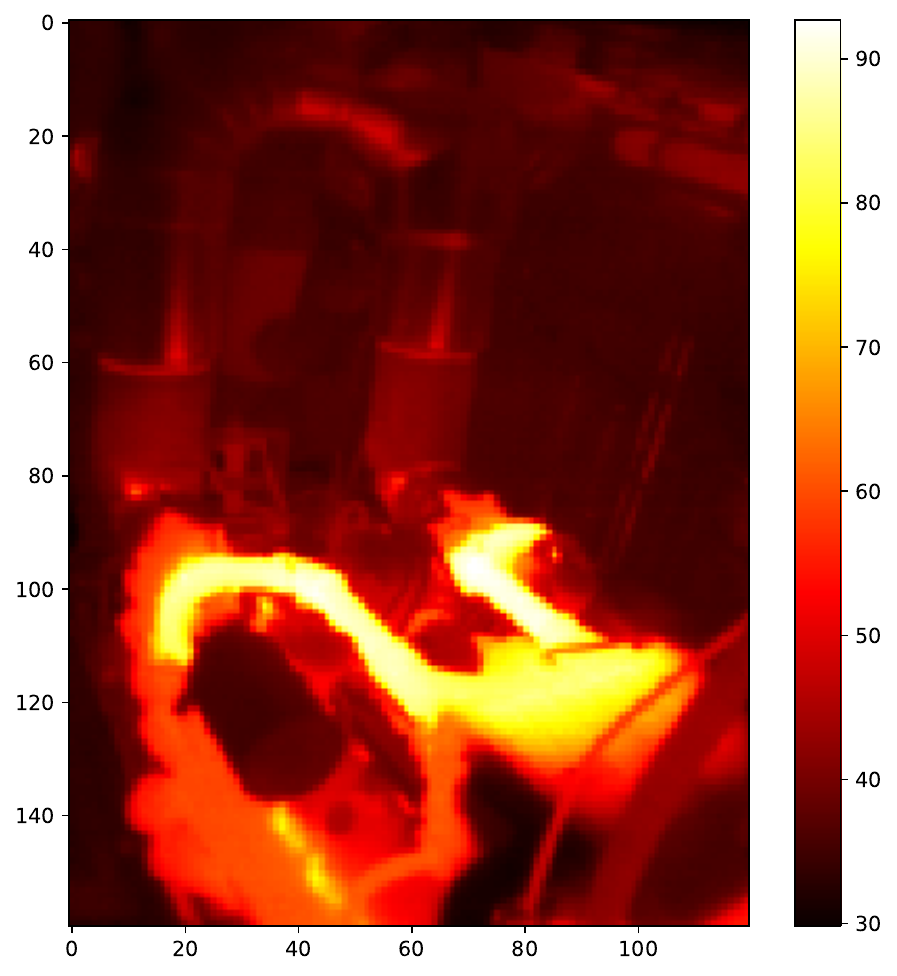}
    \caption{Unperturbed thermal camera image mapping a ship's engine.}
    \label{fig:Clean_image}
\end{figure}

The data acquisition spanned a total of four consecutive days. On each day, the engine was continuously monitored for a duration of approximately six hours, resulting in a cumulative observation period of 24 hours over the four days. However, the sampling frequency of the recordings was not consistent. The average time between consecutive samples is approximately 0.5 seconds. A snapshot of a thermal image mapping the ship's engine is depicted in Fig.~\ref{fig:Clean_image}.

Each image sourced from the thermal camera comprises 19,200 pixels, with dimensions set at 120x160 pixels. Each pixel captures the thermal radiations from the engine, which can potentially offer insights into the ship's engine's thermal performance and any anomalous patterns or hotspots that may arise during its operation. 

\subsection{Perturbations} \label{Section: Perturbations}
To evaluate the methods under various conditions, we performed four simulation scenarios comprising outliers, corruptions, noise, and a combination of them. 

%The simulations are divided into five scenarios (\textbf{S1-S4}), where the first snapshot of each simulation is depicted in Fig. \ref{fig:Scenarios}.

\subsubsection*{Scenario 1}
The data is perturbed by Gaussian noise, where the noise was generated with a mean of 0 and a standard deviation of 4, ensuring that the noise values are concentrated mainly within the range of [-4, 4]. These parameters were chosen to mimic the range of intense noise present in the measurement processes.

\subsubsection*{Scenario 2} 
The data is perturbed by outliers. These outliers were introduced by randomly selecting 100 data points (pixels) and replacing their original values with randomly generated values in the range of [30, 40] and [-40, -30]. This range was chosen to ensure that the magnitude of the outliers was significantly different from that of the actual variables to simulate severe measurement anomalies. 

\subsubsection*{Scenario 3}
The data is perturbed by corruptions. These corruptions were simulated by adding uniformly distributed random noise to 10\% of the dataset over the interval [-15, 30]. This interval was selected to ensure a substantial magnitude for the corruptions, to pretend a distortion, and to provide a stringent test for the robustness of the PCA, RPCA, and OSP algorithms. 

\subsubsection*{Scenario 4} 
The data is perturbed by a combination of the previously mentioned scenarios 1, 2, and 3, leading to a superposition of all scenarios.

\subsection{LSTM network architecture}
To train the LSTM network, various parameterizations were tested. Finally, the parameters shown in Table~\ref{tab:LSTM_parameters} were chosen. The network was trained using the Adam optimizer, where the Root-Mean-Squared Error (RMSE) was set as a metric to evaluate the model's performance during training. For the predictions, we trained the network with a window size of 50 historical samples, and the considered forecast time was chosen to be 100 time steps. The network structure consists of an input layer, an LSTM layer, a dense feedforward layer, and an output layer. Since deep neural networks with numerous parameters often overfit, we inserted a dropout layer. Dropout is a technique to address overfitting in which, during training, random units and their connections are omitted \cite{nitish_srivastava_geoffrey_hinton_alex_krizhevsky_ilya_sutskever_and_ruslan_salakhutdinov_dropout_2014}.

\begin{table}[htbp]
\caption{LSTM neural network training parameters}
\begin{center}
\renewcommand{\arraystretch}{1.5} % Adjusting the row spacing
\begin{tabular}{|c|c|}
\hline
\textbf{Parameter} & \textbf{Value} \\
\hline
\hline
Input layer shape & $50\times 10$ \\
\hline
LSTM layer size (neurons) & 128 \\
\hline
Dropout ratio & 0.2 \\
\hline
Feedforward layer size (neurons) & 128 \\
\hline
Output layer size (neurons) & 10 \\
\hline
Learning rate & $10^{-4}$ \\
\hline
Epochs & 100 \\
\hline
\end{tabular}
\label{tab:LSTM_parameters}
\end{center}
\end{table}

\section{Results and Discussion}
In the following, the results of the individual approaches regarding data cleaning, data compression, and data-driven modeling are discussed.

\subsection{Data Cleaning}
\begin{figure*}[!]
    \centering
    \begin{minipage}{0.2\linewidth}
        \centering
        \includegraphics[width=\linewidth]{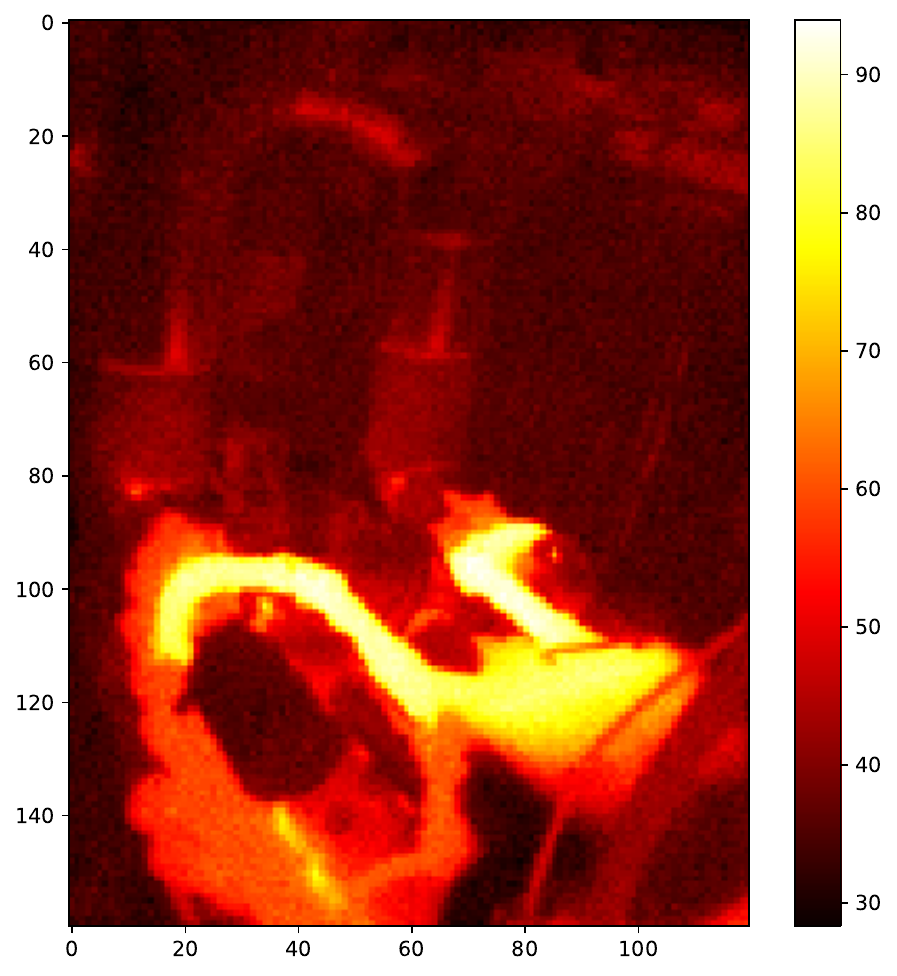}
        \\%[1ex] % Add some vertical space
        \scriptsize Scenario 1 (noise)
    \end{minipage}\hspace*{0.2cm}
    \begin{minipage}{0.2\linewidth}
        \centering
        \includegraphics[width=\linewidth]{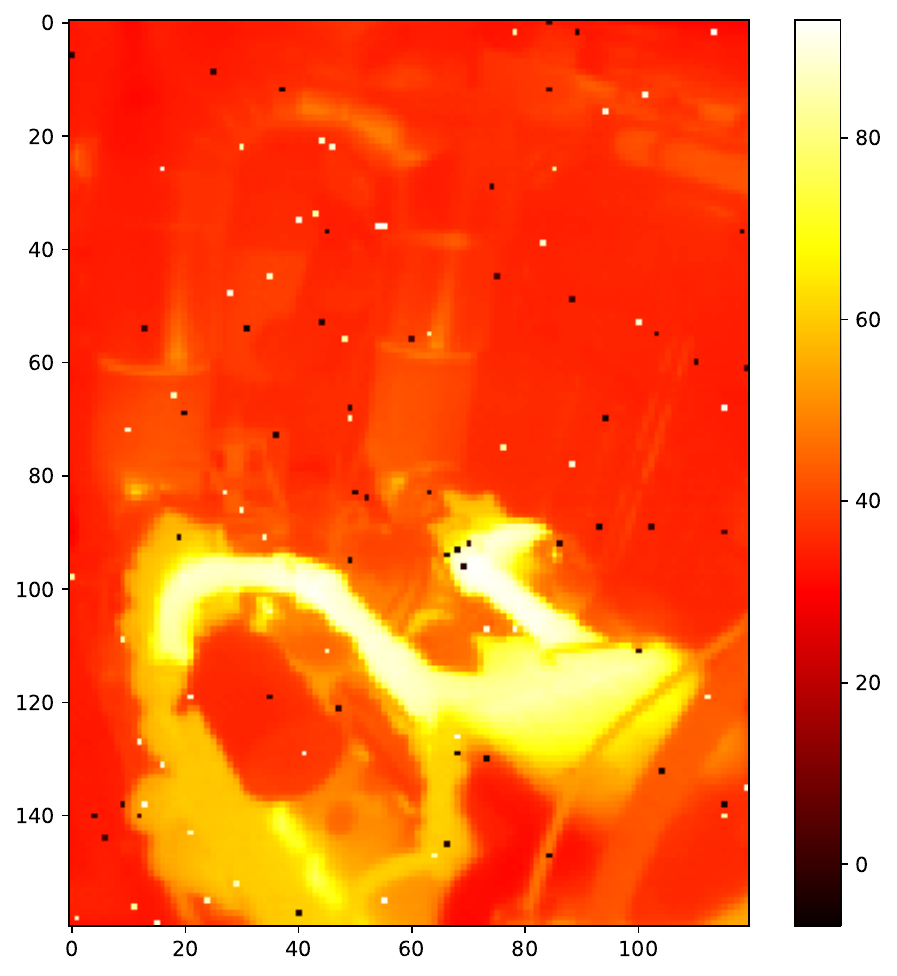}
        \\%[1ex] % Add some vertical space
        \scriptsize Scenario 2 (outliers)
    \end{minipage}\hspace*{0.2cm}
    \begin{minipage}{0.2\linewidth}
        \centering
        \includegraphics[width=\linewidth]{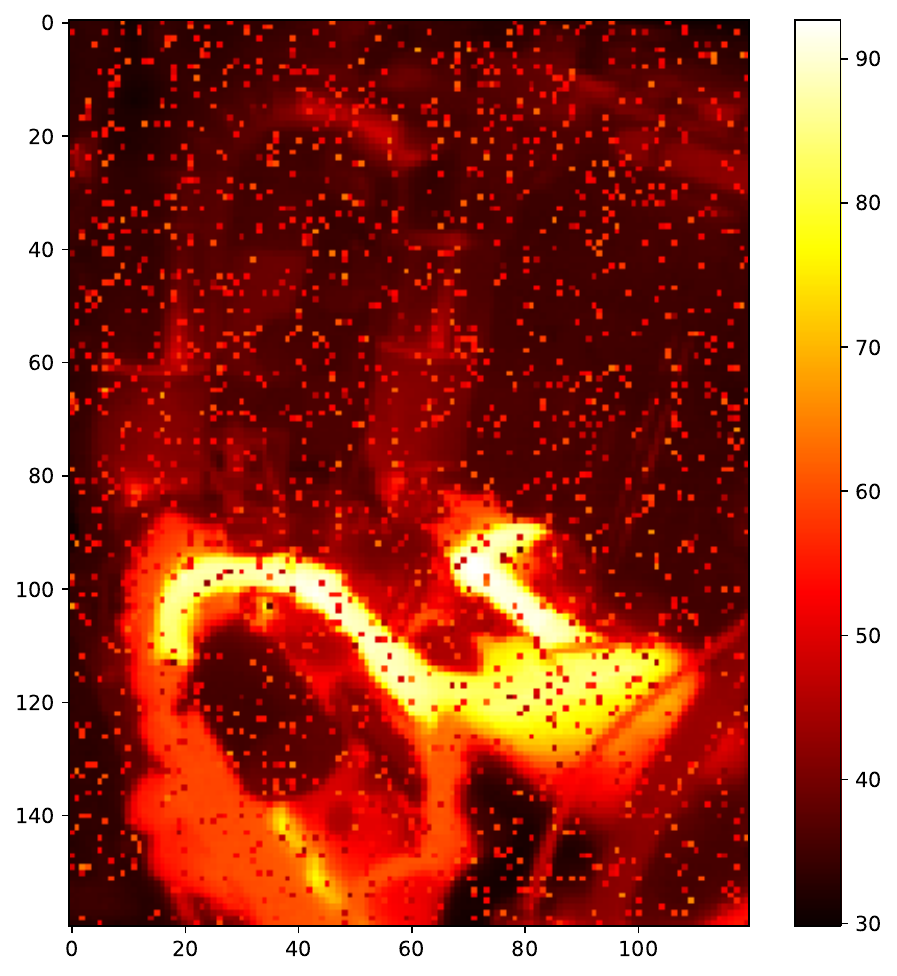}
        \\%[1ex] % Add some vertical space
        \scriptsize Scenario 3 (corruptions)
    \end{minipage}\hspace*{0.2cm}
    \begin{minipage}{0.2\linewidth}
        \centering
        \includegraphics[width=\linewidth]{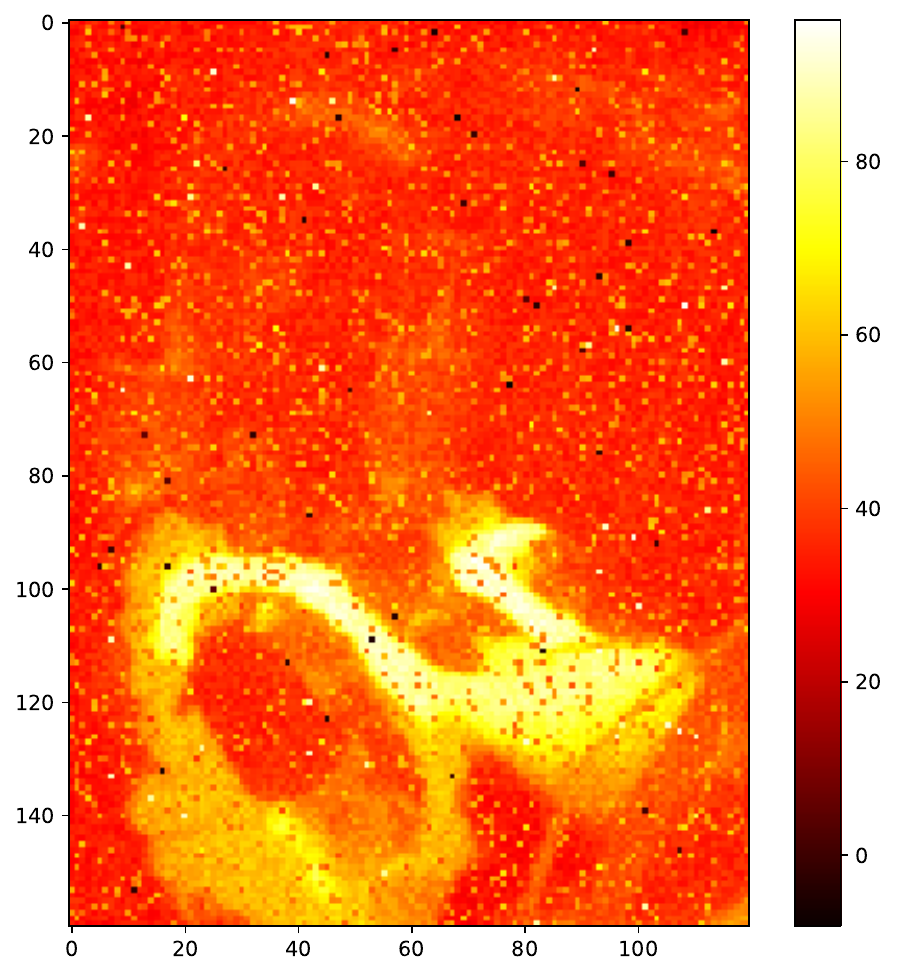}
        \\%[1ex] % Add some vertical space
        \scriptsize Scenario 4 (superposition of 1, 2, 3)
    \end{minipage}\\[2ex] % Add some vertical space

    \begin{minipage}{0.2\linewidth}
        \centering
        \includegraphics[width=\linewidth]{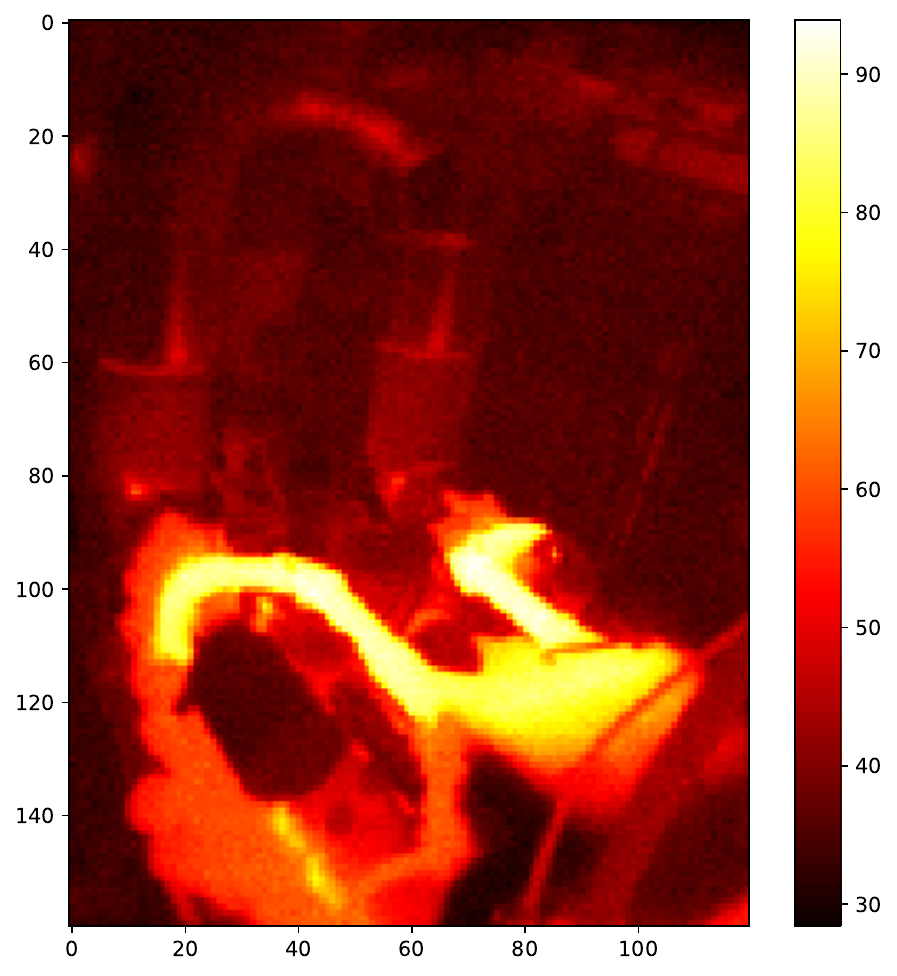}
        \\%[1ex] % Add some vertical space
        \scriptsize Reconstructed image from PCA
    \end{minipage}\hspace*{0.2cm}
    \begin{minipage}{0.2\linewidth}
        \centering
        \includegraphics[width=\linewidth]{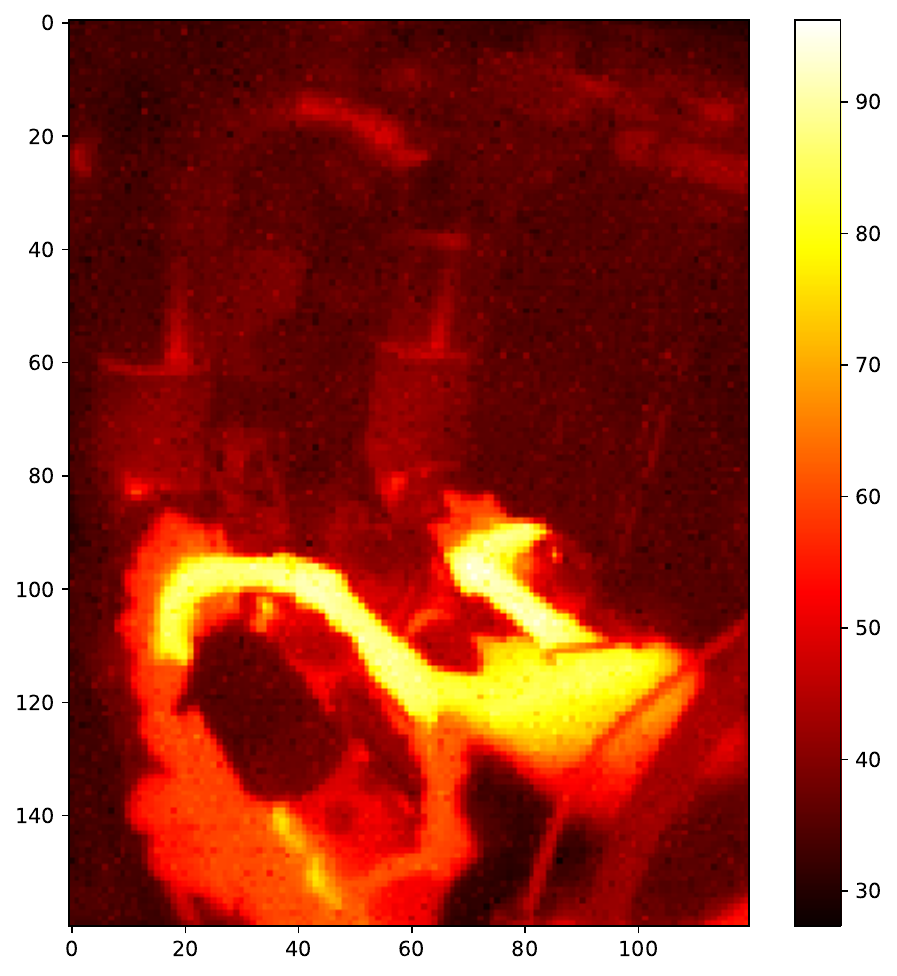}
        \\%[1ex] % Add some vertical space
        \scriptsize Reconstructed image from PCA
    \end{minipage}\hspace*{0.2cm}
    \begin{minipage}{0.2\linewidth}
        \centering
        \includegraphics[width=\linewidth]{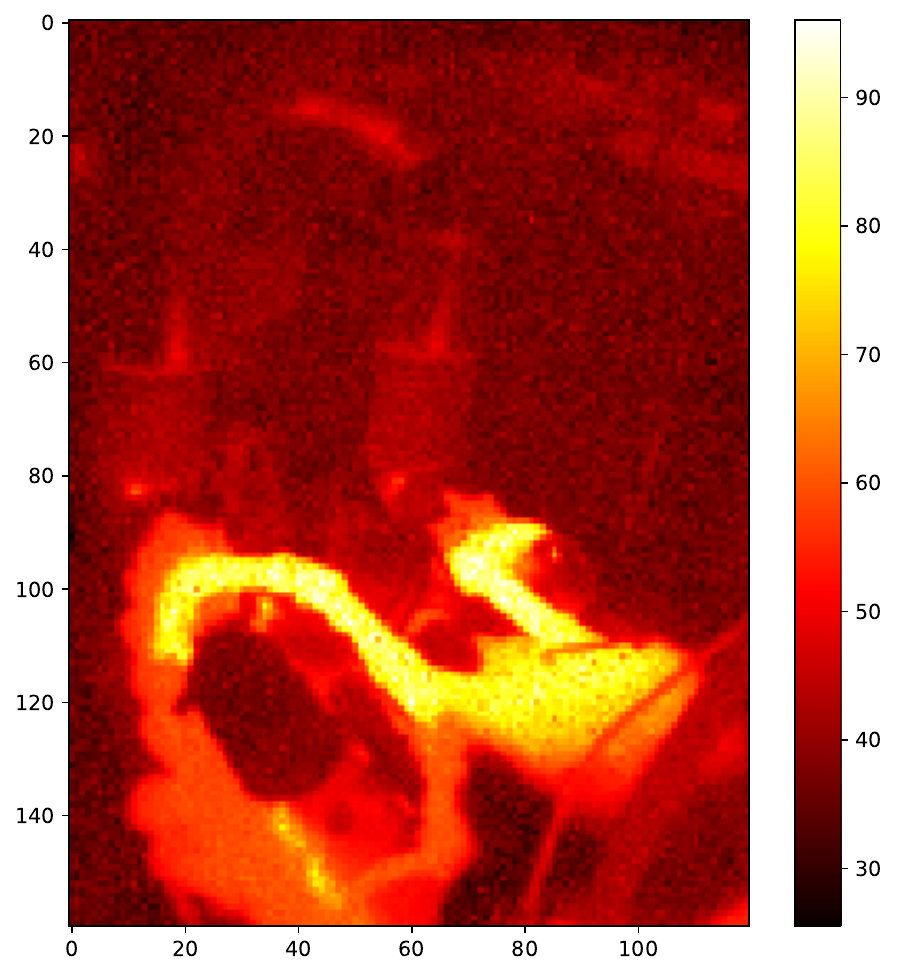}
        \\%[1ex] % Add some vertical space
        \scriptsize Reconstructed image from PCA
    \end{minipage}\hspace*{0.2cm}
    \begin{minipage}{0.2\linewidth}
        \centering
        \includegraphics[width=\linewidth]{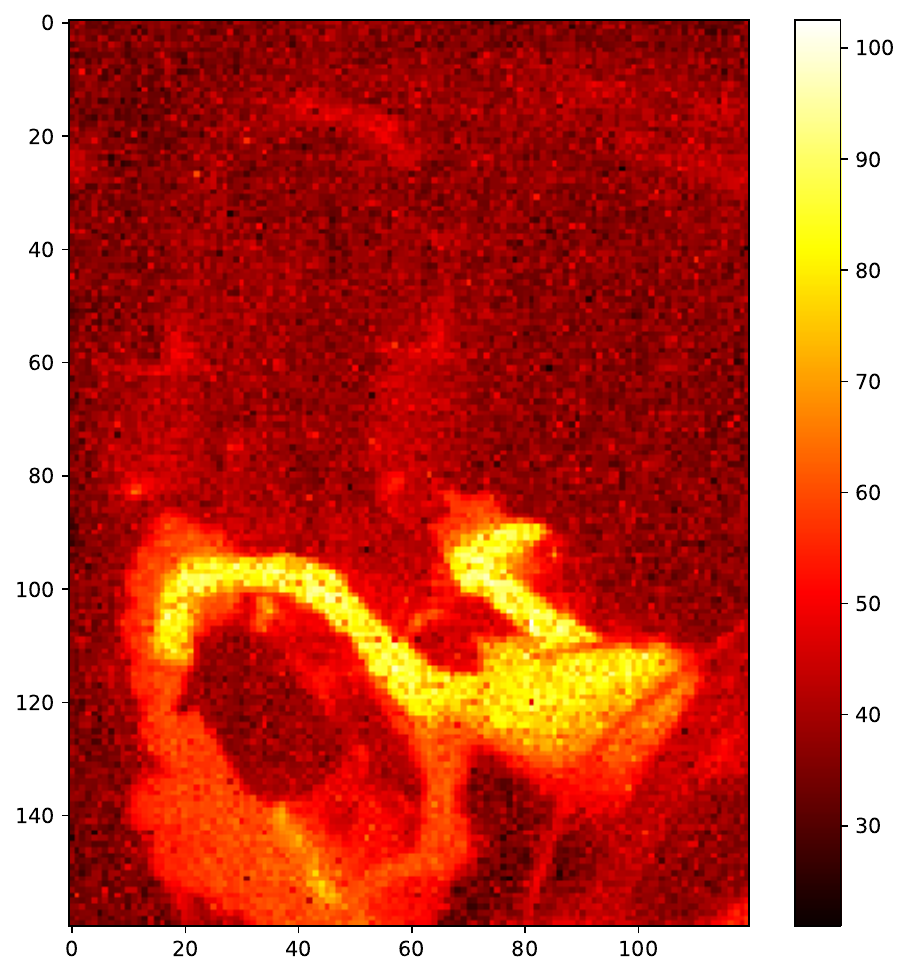}
        \\%[1ex] % Add some vertical space
        \scriptsize Reconstructed image from PCA
    \end{minipage}\\[2ex] % Add some vertical space

    \begin{minipage}{0.2\linewidth}
        \centering
        \includegraphics[width=\linewidth]{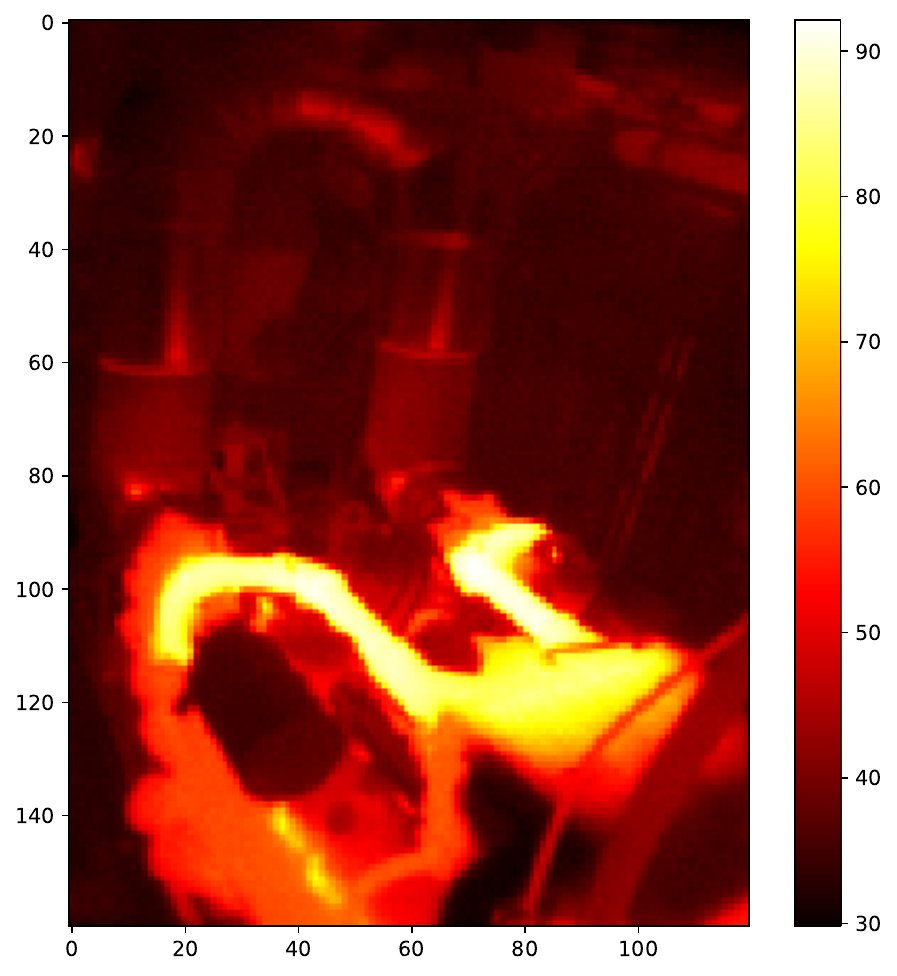}
        \\%[1ex] % Add some vertical space
        \scriptsize Matrix $\mathbf{L}$ from RPCA
    \end{minipage}\hspace*{0.2cm}
    \begin{minipage}{0.2\linewidth}
        \centering
        \includegraphics[width=\linewidth]{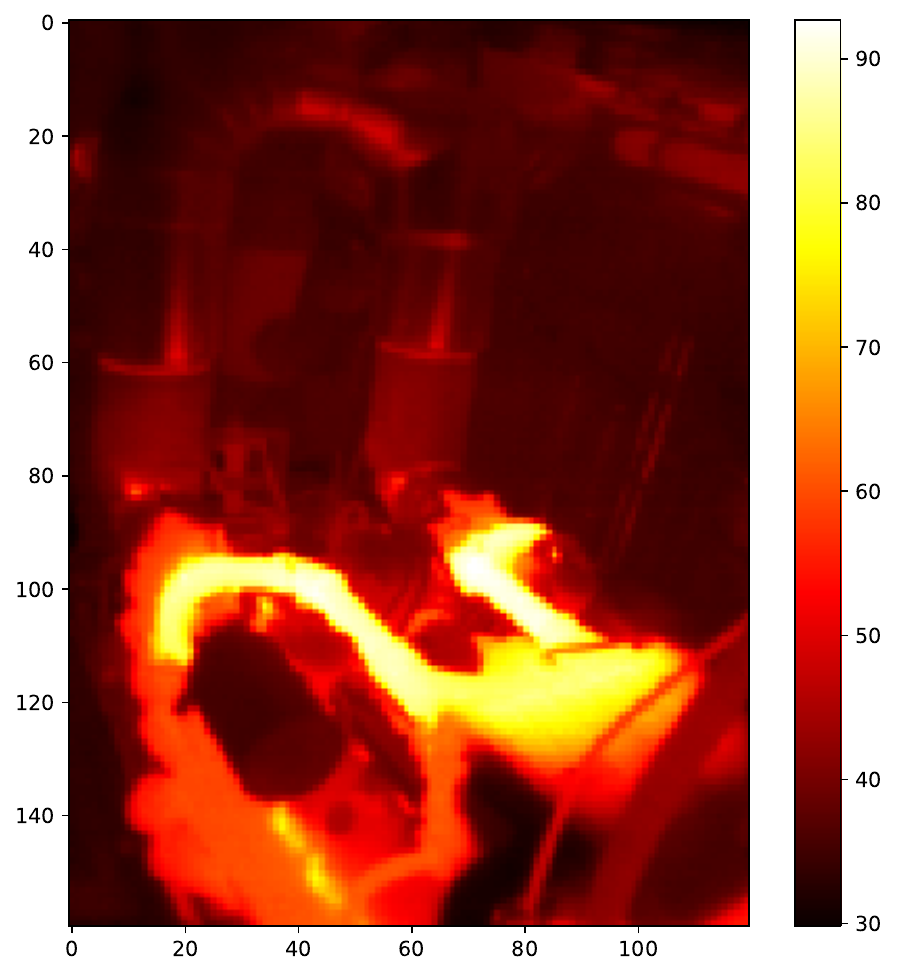}
        \\%[1ex] % Add some vertical space
        \scriptsize Matrix $\mathbf{L}$ from RPCA
    \end{minipage}\hspace*{0.2cm}
    \begin{minipage}{0.2\linewidth}
        \centering
        \includegraphics[width=\linewidth]{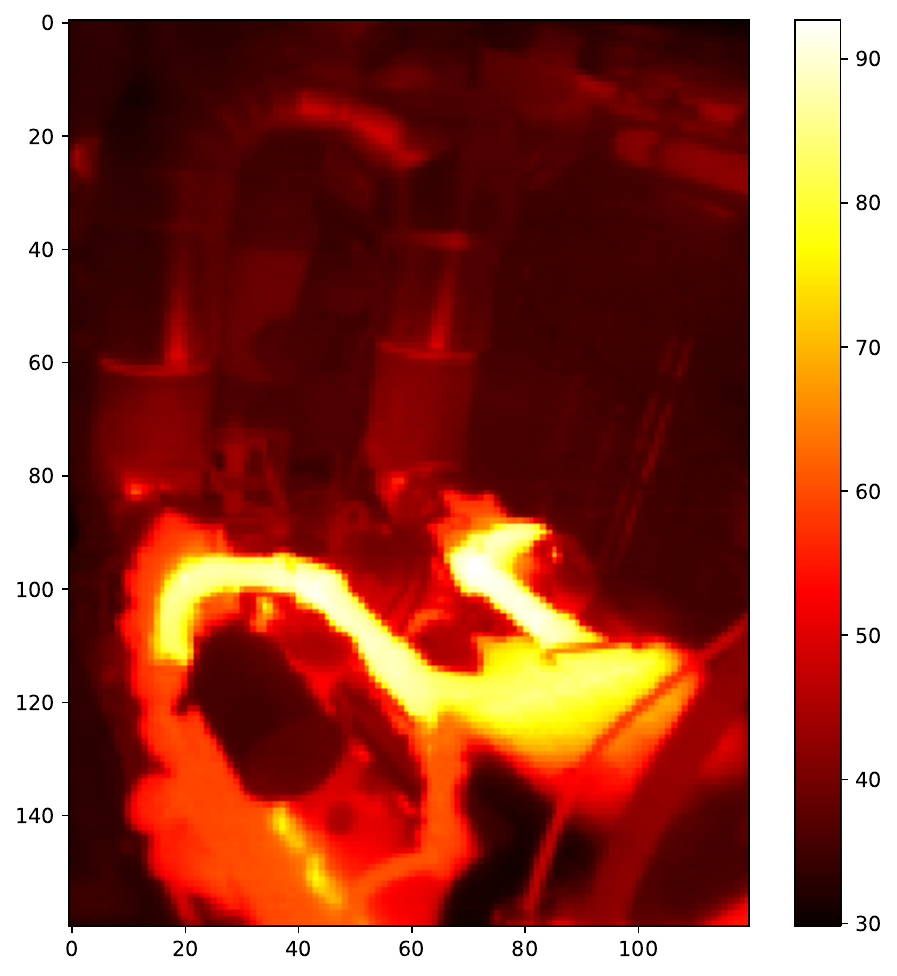}
        \\%[1ex] % Add some vertical space
        \scriptsize Matrix $\mathbf{L}$ from RPCA
    \end{minipage}\hspace*{0.2cm}
    \begin{minipage}{0.2\linewidth}
        \centering
        \includegraphics[width=\linewidth]{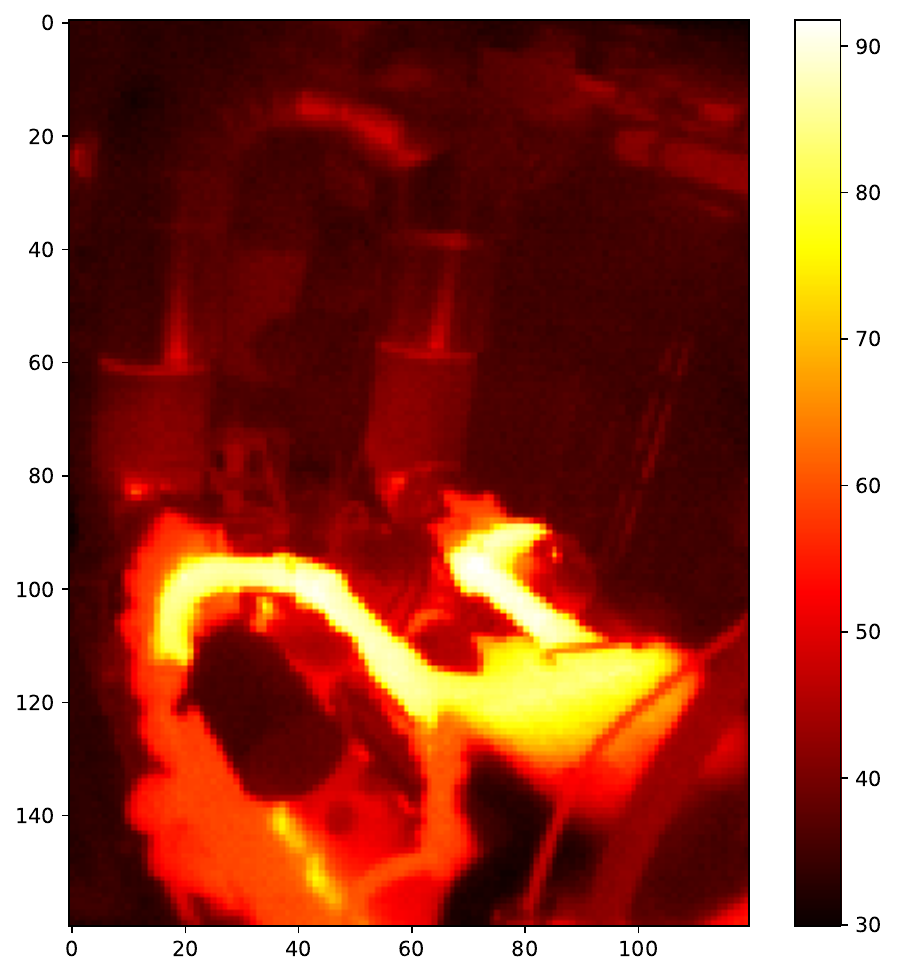}
        \\%[1ex] % Add some vertical space
        \scriptsize Matrix $\mathbf{L}$ from RPCA
    \end{minipage}\\[2ex] % Add some vertical space
    
    \begin{minipage}{0.2\linewidth}
        \centering
        \includegraphics[width=\linewidth]{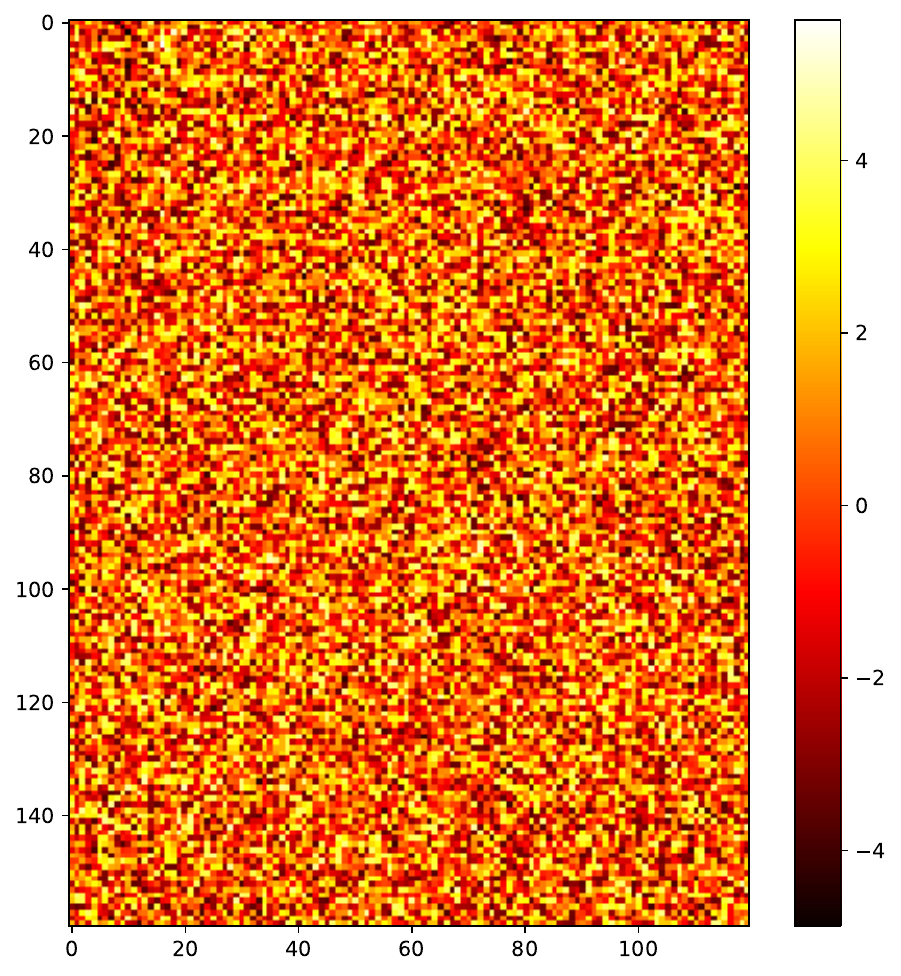}
        \\%[1ex] % Add some vertical space
        \scriptsize Matrix $\mathbf{S}$ from RPCA
    \end{minipage}\hspace*{0.2cm}
    \begin{minipage}{0.2\linewidth}
        \centering
        \includegraphics[width=\linewidth]{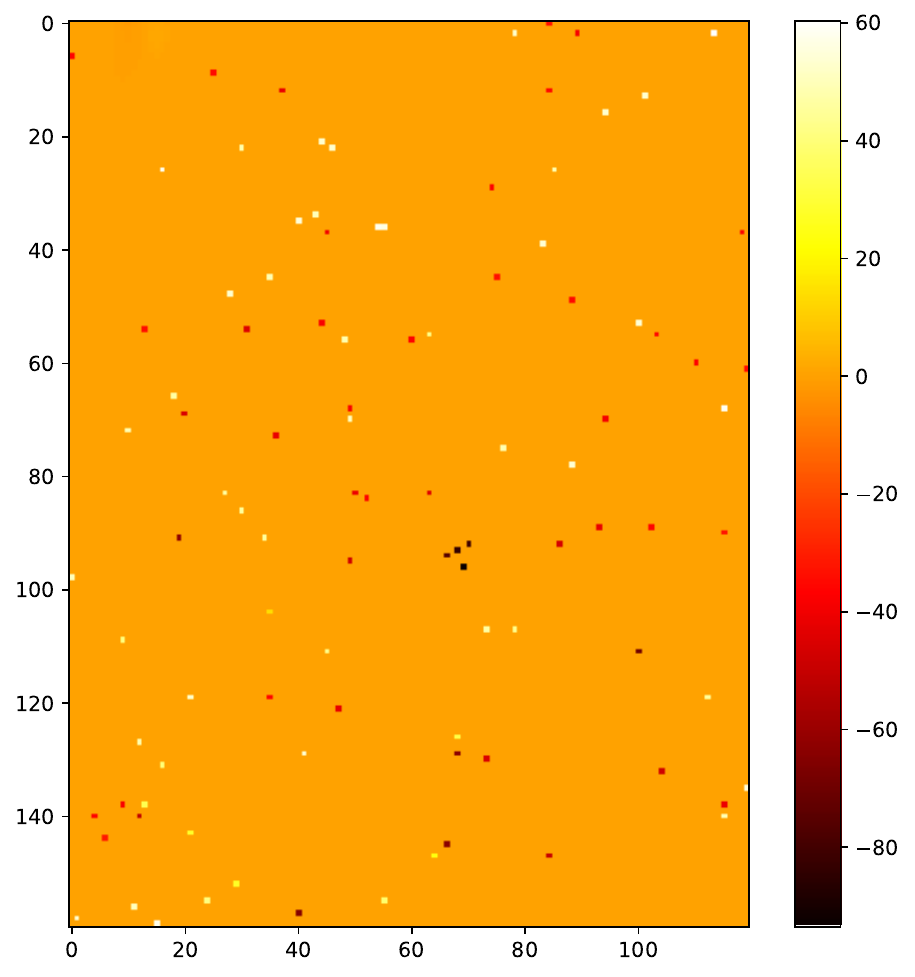}
        \\%[1ex] % Add some vertical space
        \scriptsize Matrix $\mathbf{S}$ from RPCA
    \end{minipage}\hspace*{0.2cm}
    \begin{minipage}{0.2\linewidth}
        \centering
        \includegraphics[width=\linewidth]{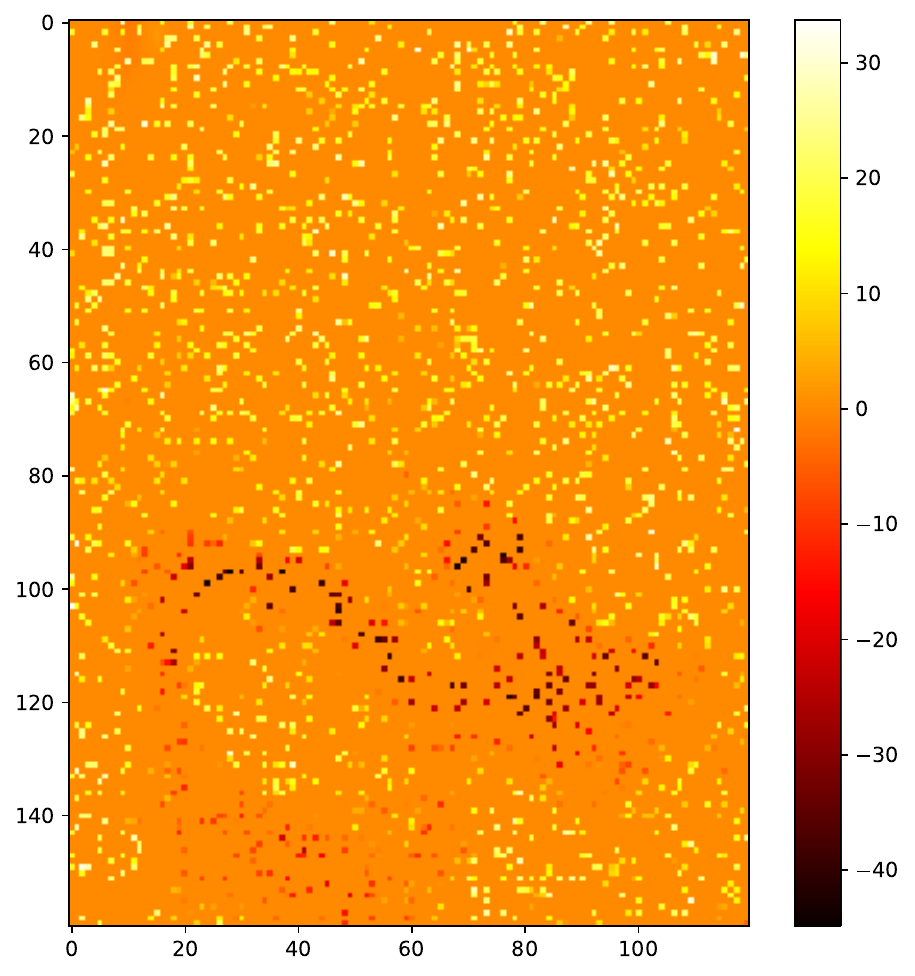}
        \\%[1ex] % Add some vertical space
        \scriptsize Matrix $\mathbf{S}$ from RPCA
    \end{minipage}\hspace*{0.2cm}
    \begin{minipage}{0.2\linewidth}
        \centering
        \includegraphics[width=\linewidth]{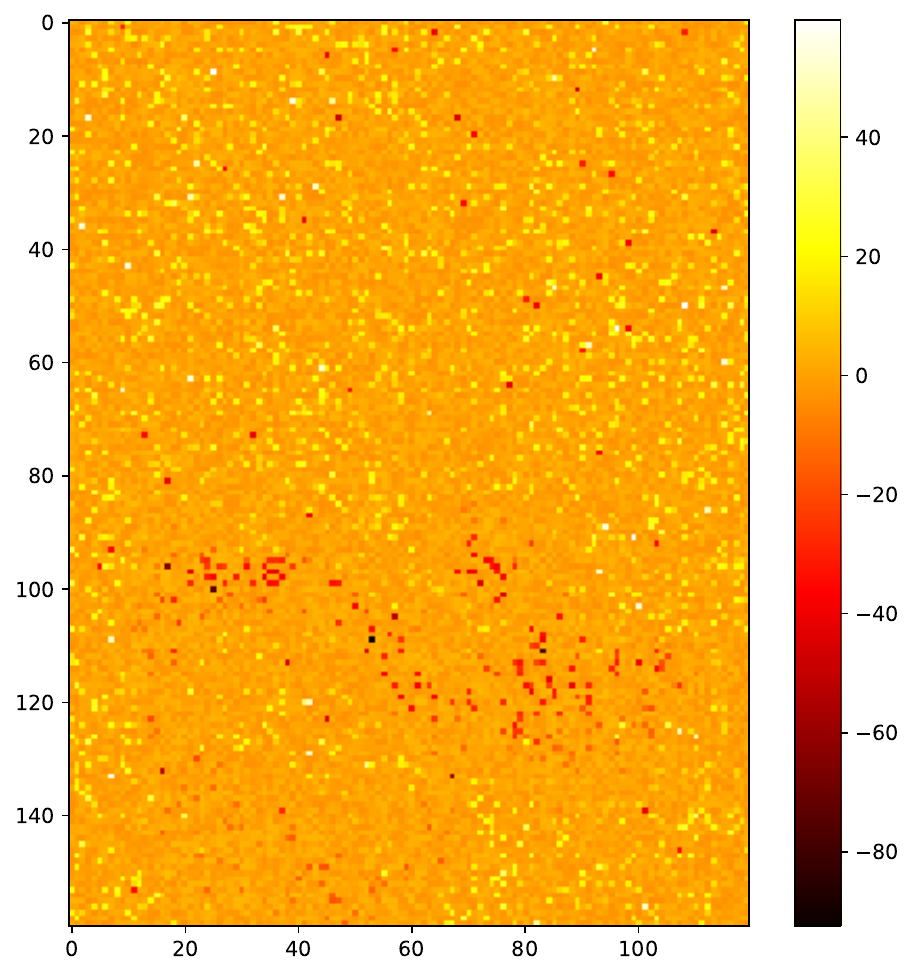}
        \scriptsize Matrix $\mathbf{S}$ from RPCA
        \\%[1ex] % Add some vertical space
    \end{minipage}\\[1ex] % Add some vertical space
    \caption{Results of RPCA and PCA applied on thermal camera data under various conditions. The different scenarios are described in Section \ref{Section: Perturbations}.}
    \label{fig:RPCA_results}
\end{figure*}

The data cleaning phase is demonstrated in Fig.~\ref{fig:RPCA_results}. Presented are the four various scenarios described in Section~\ref{Section: Perturbations}. Note that the unperturbed image of Fig.~\ref{fig:Clean_image} reflects the ground truth. The results of RPCA are compared with those of PCA. It is shown how RPCA decomposes the thermal image data into the matrices $\mathbf{L}$ and $\mathbf{S}$. 
The matrix $\mathbf{L}$ clearly depicts the unperturbed image, while the matrix $\mathbf{S}$ captures the sparse components of the data, which mainly contain all unwanted fragments and anomalies.
In contrast, the image reconstructions of traditional PCA are especially susceptible to intensive corruptions and outliers. 
Therefore, the capability of RPCA to decompose data into a low-rank matrix $\mathbf{L}$ and a sparse matrix $\mathbf{S}$ can improve the accuracy of many AI applications utilizing big data.

\subsection{Data Compression}

\begin{figure*}[htb!]
    \centering
    \begin{minipage}{0.3\linewidth}
        \centering
        \includegraphics[width=\linewidth]{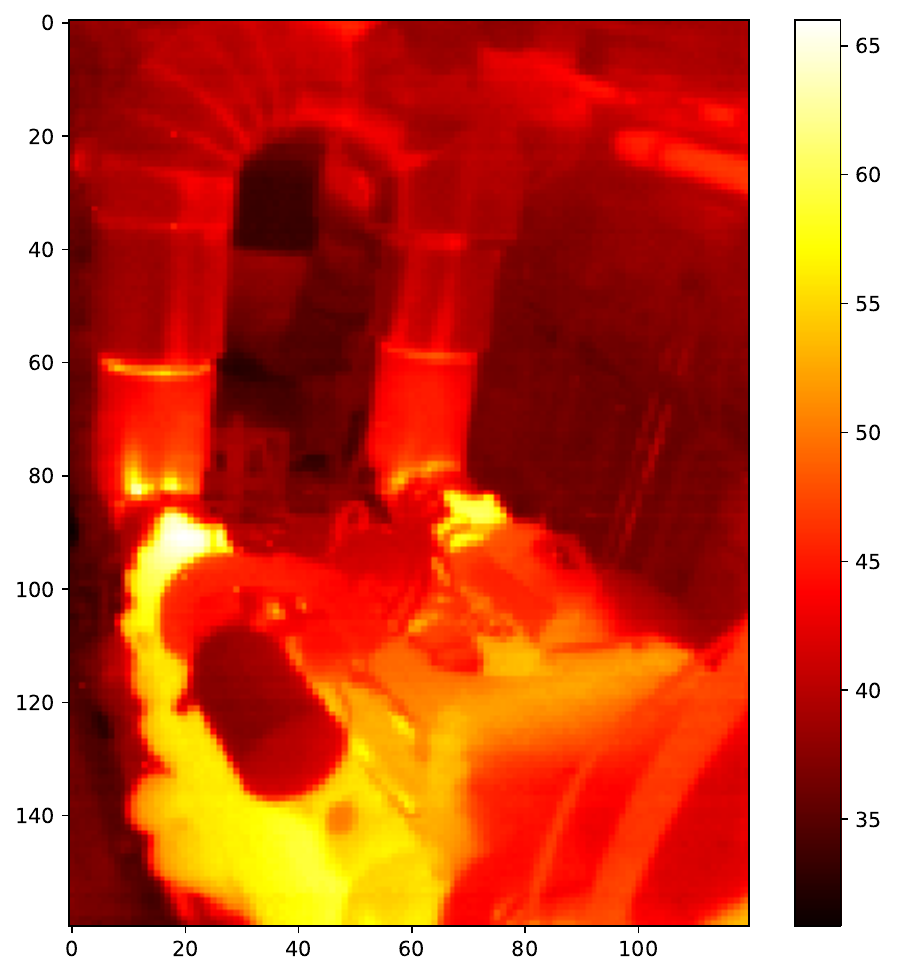}
        \\[1ex] % Add some vertical space
        \small Original image
    \end{minipage}\hspace*{0.5cm}
    \begin{minipage}{0.3\linewidth}
        \centering
        \includegraphics[width=\linewidth]{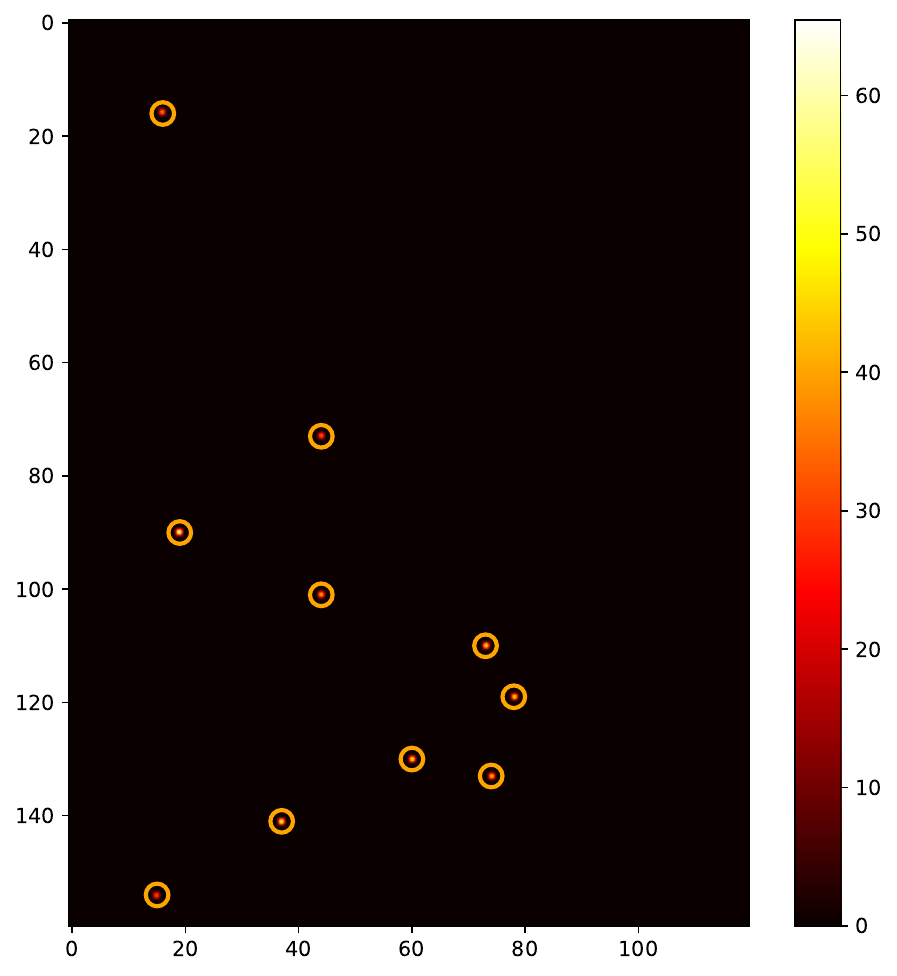}
        \\[1ex] % Add some vertical space
        \small Measurements
    \end{minipage}\hspace*{0.5cm}
    \begin{minipage}{0.3\linewidth}
        \centering
        \includegraphics[width=\linewidth]{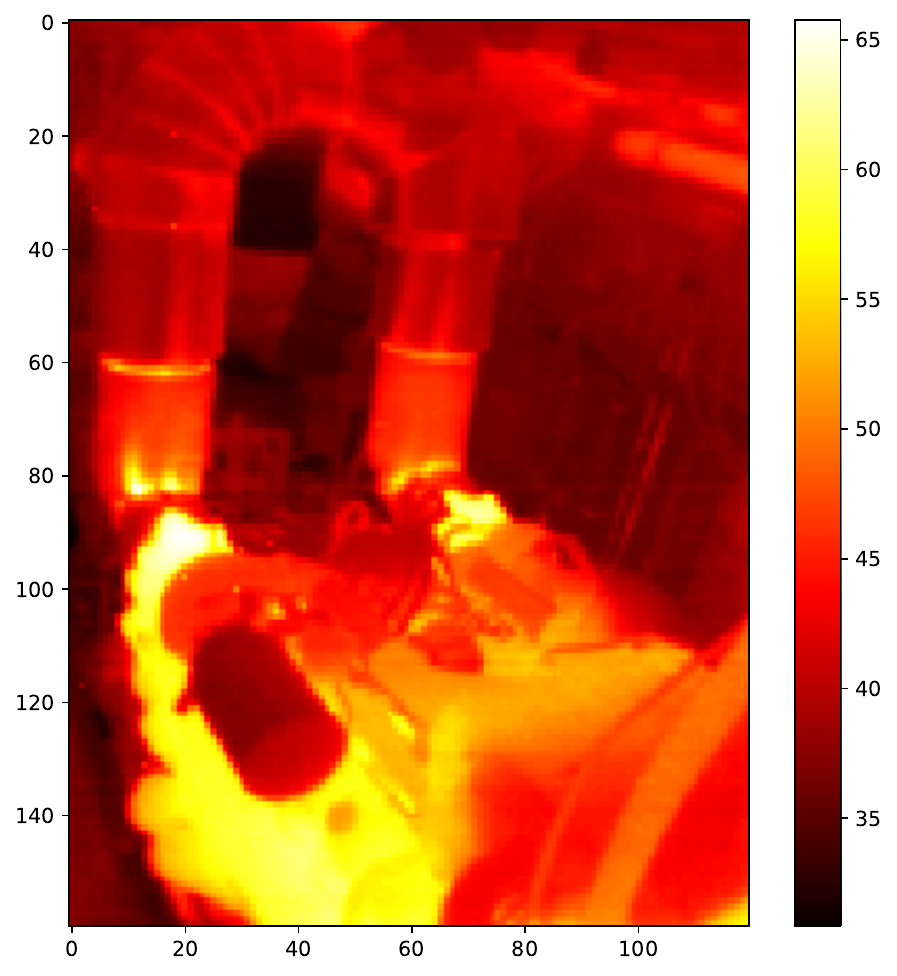}
        \small Reconstructed image
        \\[1ex] % Add some vertical space
    \end{minipage}\\[1ex] % Add some vertical space
    \caption{Optimal sensor positions and their capability to reconstruct the original images.}
    \label{fig:OSP_results}
\end{figure*}

OSP applied to the thermal image data can drastically reduce the data dimension. In this study, we used only 10 of the original 19,200 pixel measurements. As illustrated in Fig.~\ref{fig:OSP_results}, it is evident that the original thermal images could be accurately reconstructed using a substantially reduced set of pixel measurements.

From a data compression standpoint, the ability to reconstruct comprehensive thermal images using limited pixel measurements underscores the energy and memory efficiency of OSP. This reduced representation not only implies a significant reduction in data size but also means that the essential features and characteristics of the thermal images are captured with minimal loss of information. Consequently, this data compression approach enables faster processing times, reduced memory requirements, and lower energy usage in real-time applications or scenarios with bandwidth constraints.

Memory savings can be considered as follows. Assuming a data matrix $\mathbf{X}\in \mathbb{R}^{m\times n}$, where $m$ is the spatial dimension and $n$ expresses the time dimension, while the lower-dimensional measurement matrix $\mathbf{Y}\in \mathbb{R}^{r\times n}$ is spanned by a low dimension $r$, then the ratio of saving memory is given by
\begin{equation}
    \alpha = \frac{m}{r}.
\end{equation}
In this case study, the ratio of saving memory yields
\begin{equation}
    \alpha = \frac{19200}{10} = 1920.
\end{equation}
This implies that, under the consideration of equal memory, we can store 1920 times more thermal images.

\subsection{Predictive Data-Driven Modeling}
\begin{figure}[b]
    \centering
    \includegraphics[width=\linewidth]{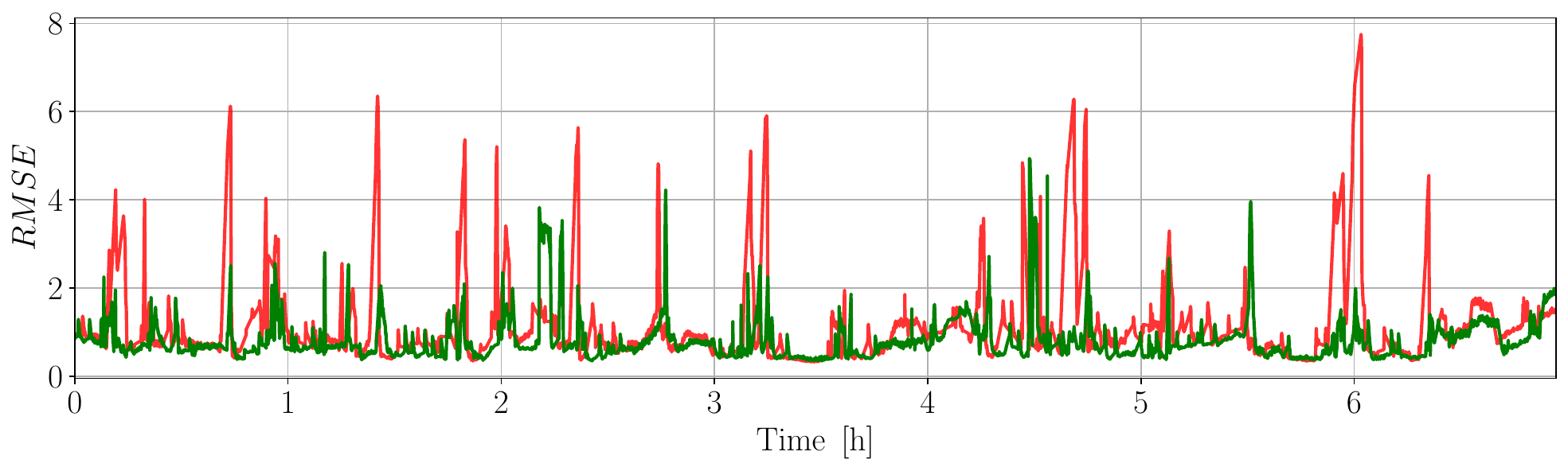}
    \caption{RMSE of the LSTM predictions after reconstructing the entire images with the few pixel measurements from OSP. Compared are the predictions for 100 timesteps, where the LSTM model is trained on the original OSP pixel data (\protect\redline) and the interpolated OSP pixel data (\protect\greenline).}
    \label{fig:LSTM_prediction_RMSE_images}
\end{figure}

The LSTM network was trained using a sparse subspace $\mathbf{Y}$, obtained by OSP. Since this study dealt with data containing inconsistent time samples, we interpolated the data before building data-driven models via LSTM networks. To show the influence of previously interpolating the data, we depict the RMSE of the model predictions concerning the data-driven model with and without an initial interpolation in Fig.~\ref{fig:LSTM_prediction_RMSE_images}. 
The RMSE is related to the reconstructed images of the original image size (19,200 pixels) using the model predictions from the few OSP measurements (10 pixels). 
\begin{figure}[b]
    \centering
    \includegraphics[width=\linewidth]{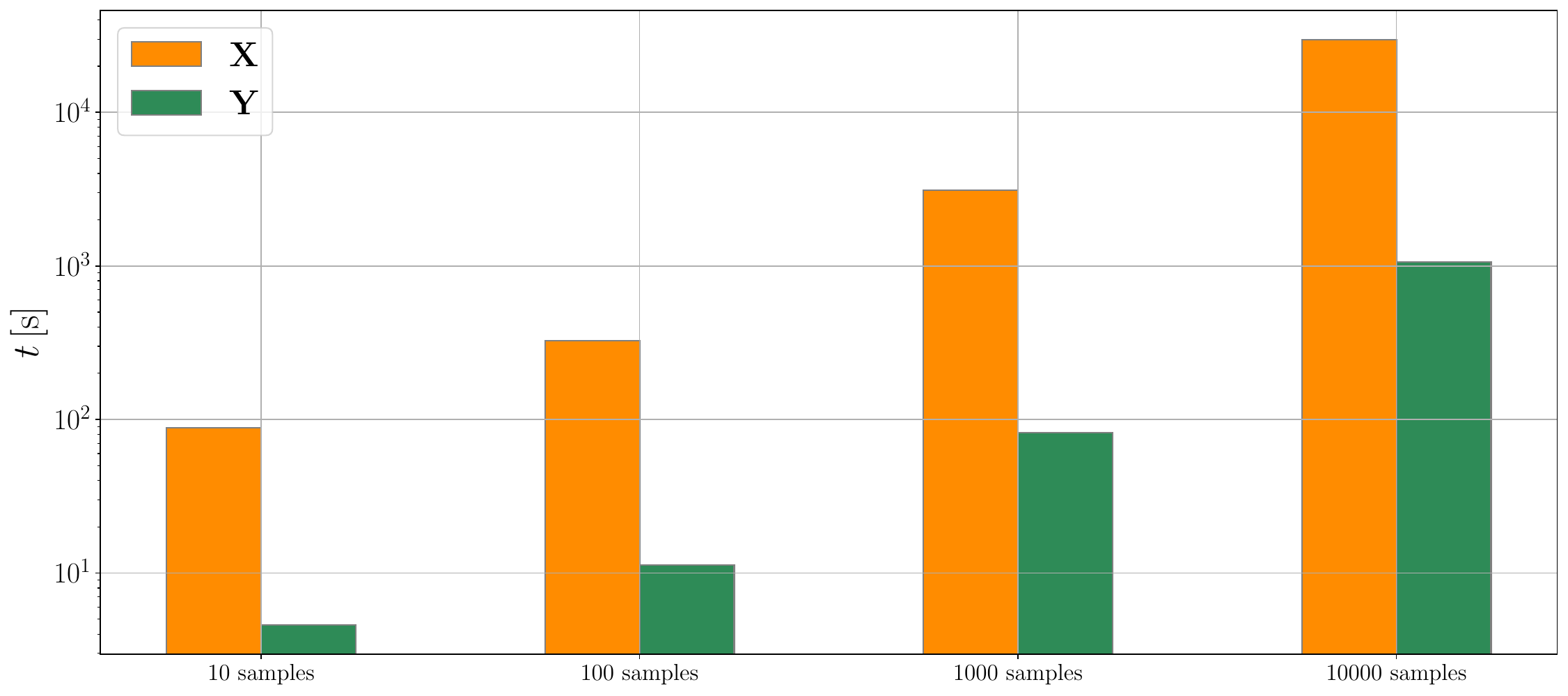}
    \caption{Comparison of the computational time for a training phase of the LSTM network regarding full image data $\mathbf{X}$ and the compressed image data $\mathbf{Y}$. The training time is compared to various data sizes. Note that the time is scaled logarithmically. The number of epochs was set relatively high (100). Therefore, changing training parameters can allow online training with a duration of less than a second.}
    \label{fig:computational_costs}
\end{figure}
Furthermore, a comparison of the computational time for a training phase is demonstrated in Fig.~\ref{fig:computational_costs}, showing the tremendous efficiency improvement of the proposed approach. For comparison, the network structure and training parameters of Table~\ref{tab:LSTM_parameters} were used. 
%It should be noted that the LSTM training process was particularly efficient and executable in just a few minutes, while training on the entire image data would have taken multiple days. 
The computational efficiency underscores the practicality of the method, especially when considering real-time applications. Once trained, the model's ability to make predictions is instantaneous, allowing real-time forecasts to be made in milliseconds. In addition, depending on the application and the parameters chosen for training (e.g., number of epochs), the proposed approach can enable online training in real-time. 
%For this purpose, the number of training epochs plays an important role in accelerating the process.

\begin{figure*}[h]
    \centering
    \begin{minipage}{0.5\linewidth}
        \centering
        \includegraphics[width=\linewidth]{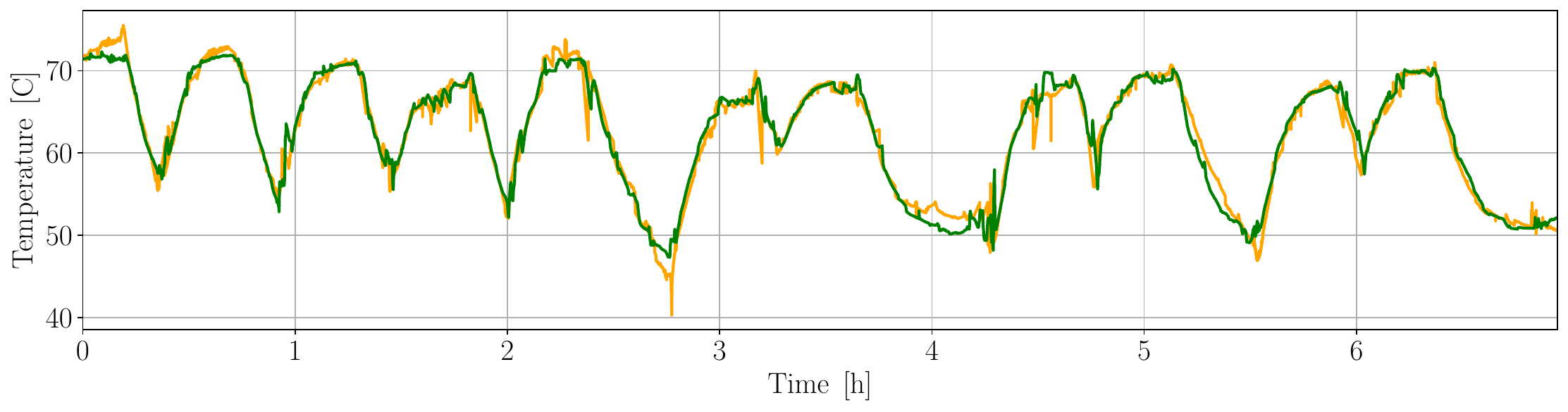}
        \\[1ex] % Add some vertical space
    \end{minipage}\hfill
    \begin{minipage}{0.5\linewidth}
        \centering
        \includegraphics[width=\linewidth]{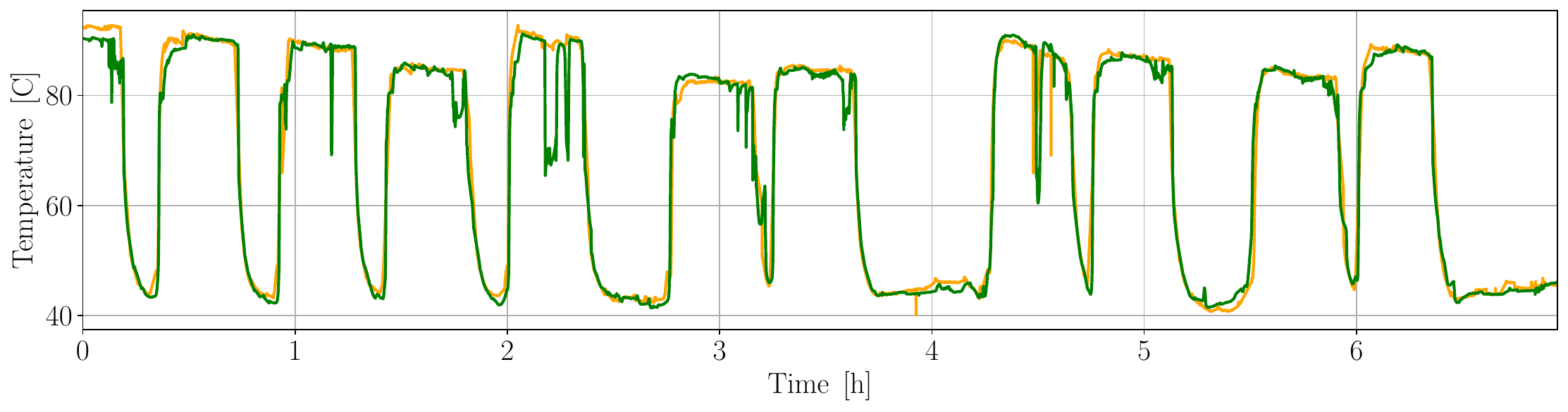}
        \\[1ex] % Add some vertical space
    \end{minipage}\\[0ex] % Add some vertical space

    \begin{minipage}{0.5\linewidth}
        \centering
        \includegraphics[width=\linewidth]{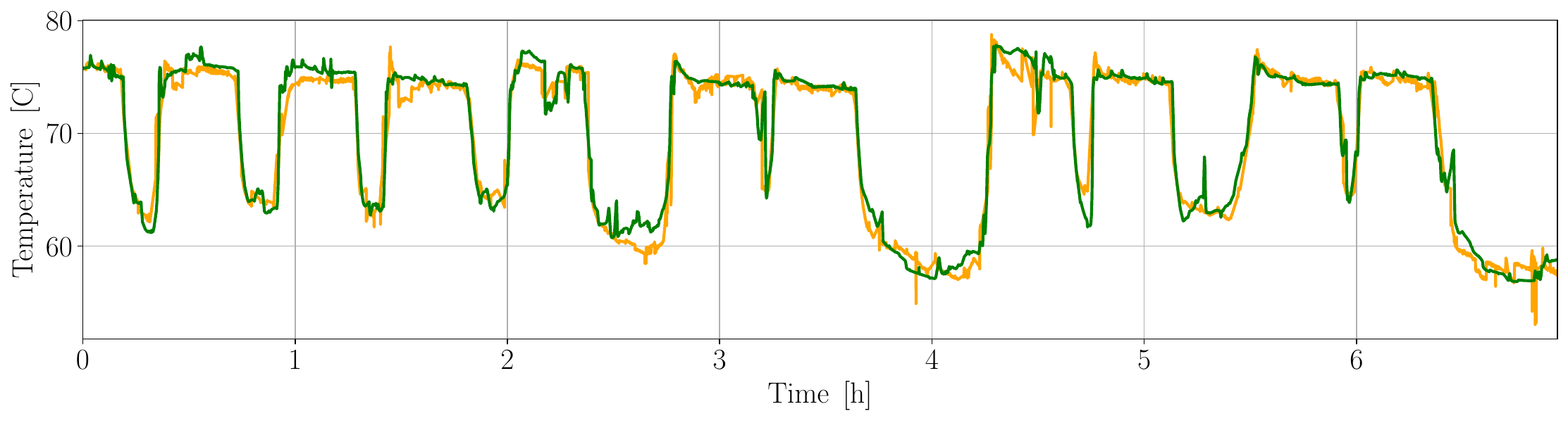}
        \\[1ex] % Add some vertical space
    \end{minipage}\hfill
    \begin{minipage}{0.5\linewidth}
        \centering
        \includegraphics[width=\linewidth]{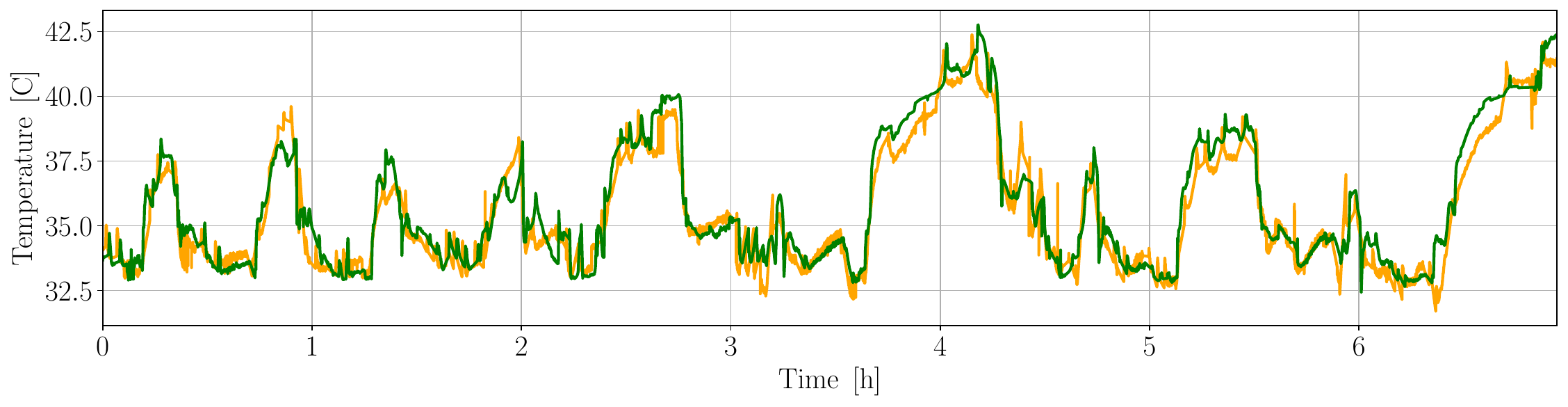}
        \\[1ex] % Add some vertical space
    \end{minipage}\\[0ex] % Add some vertical space

    \begin{minipage}{0.5\linewidth}
        \centering
        \includegraphics[width=\linewidth]{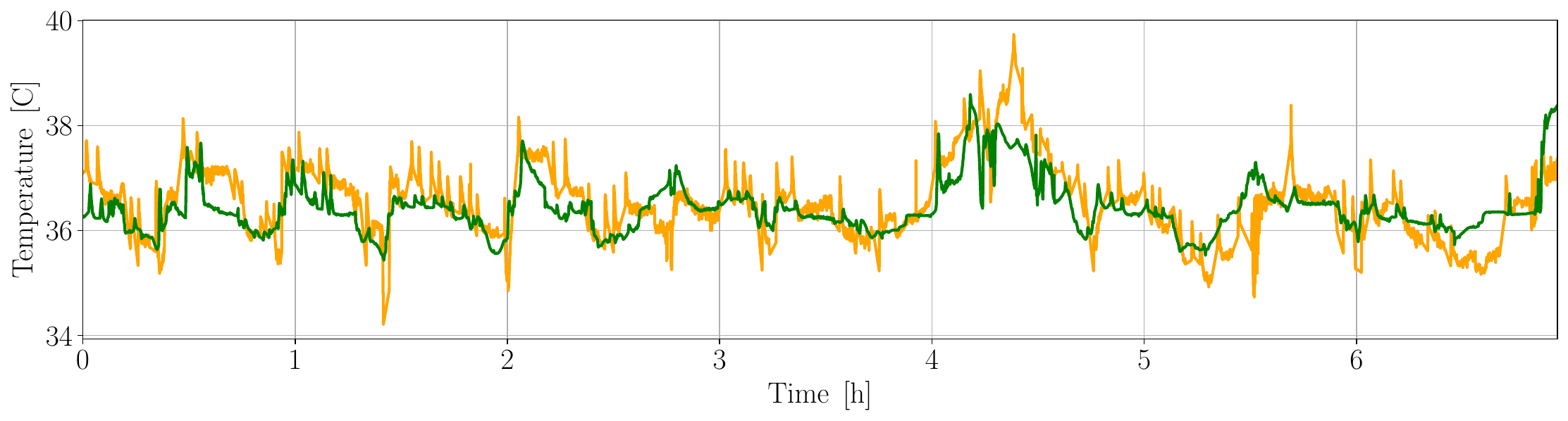}
        \\[1ex] % Add some vertical space
    \end{minipage}\hfill
    \begin{minipage}{0.5\linewidth}
        \centering
        \includegraphics[width=\linewidth]{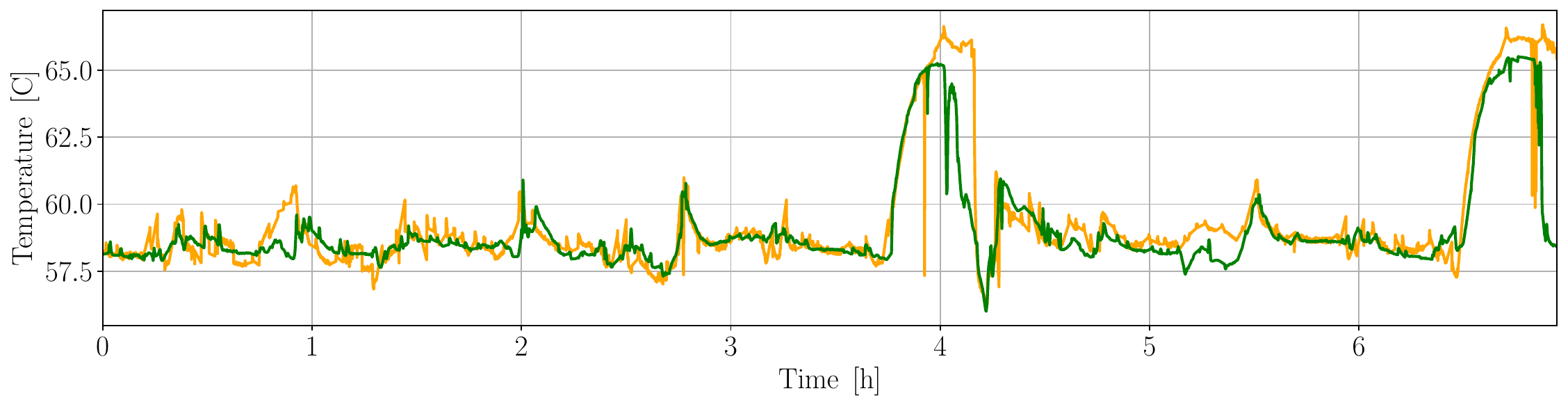}
        \\[1ex] % Add some vertical space
    \end{minipage}\\[0ex] % Add some vertical space

    \begin{minipage}{0.5\linewidth}
        \centering
        \includegraphics[width=\linewidth]{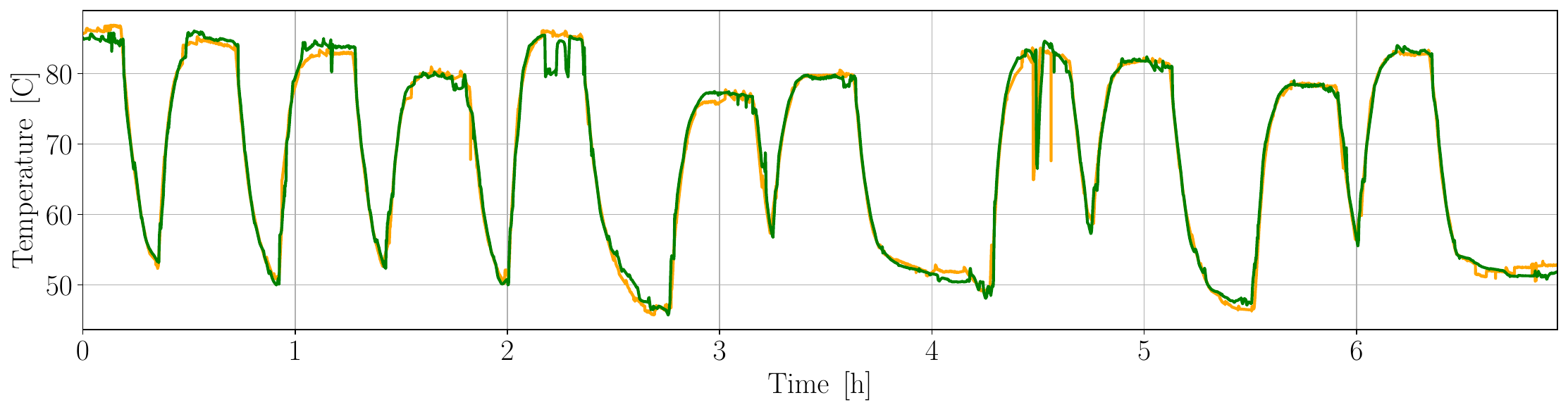}
        \\[1ex] % Add some vertical space
    \end{minipage}\hfill
    \begin{minipage}{0.5\linewidth}
        \centering
        \includegraphics[width=\linewidth]{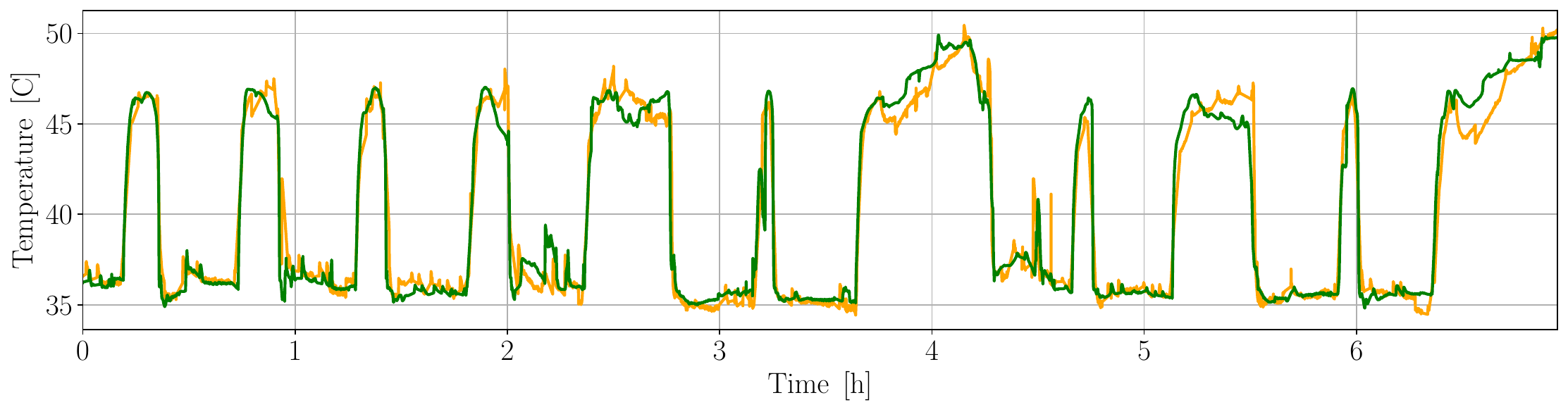}
        \\[1ex] % Add some vertical space
    \end{minipage}\\[0ex] % Add some vertical space
    
    \begin{minipage}{0.5\linewidth}
        \centering
        \includegraphics[width=\linewidth]{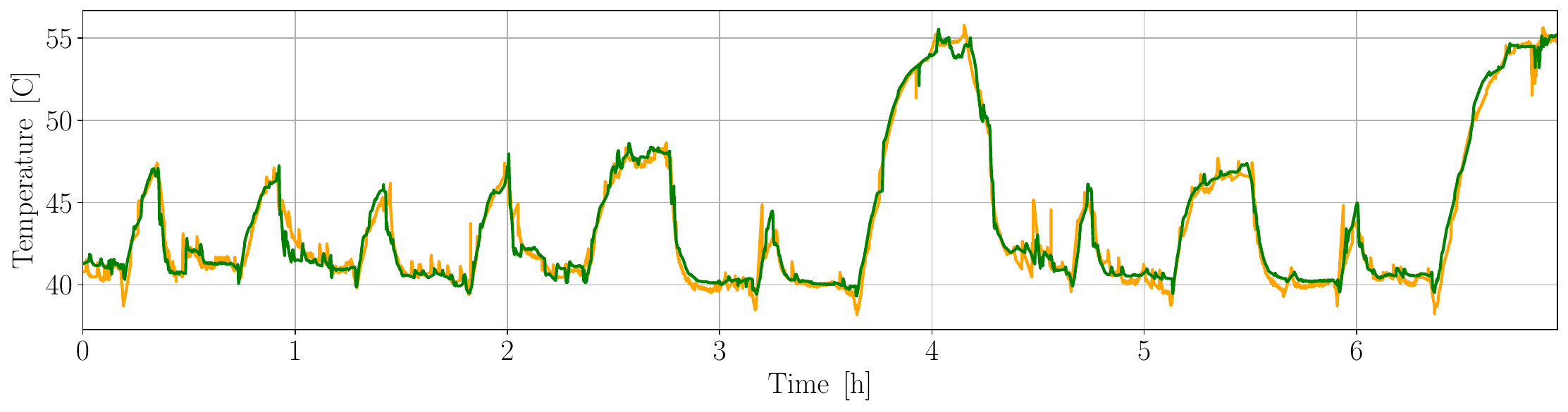}
        \\[1ex] % Add some vertical space
    \end{minipage}\hfill
    \begin{minipage}{0.5\linewidth}
        \centering
        \includegraphics[width=\linewidth]{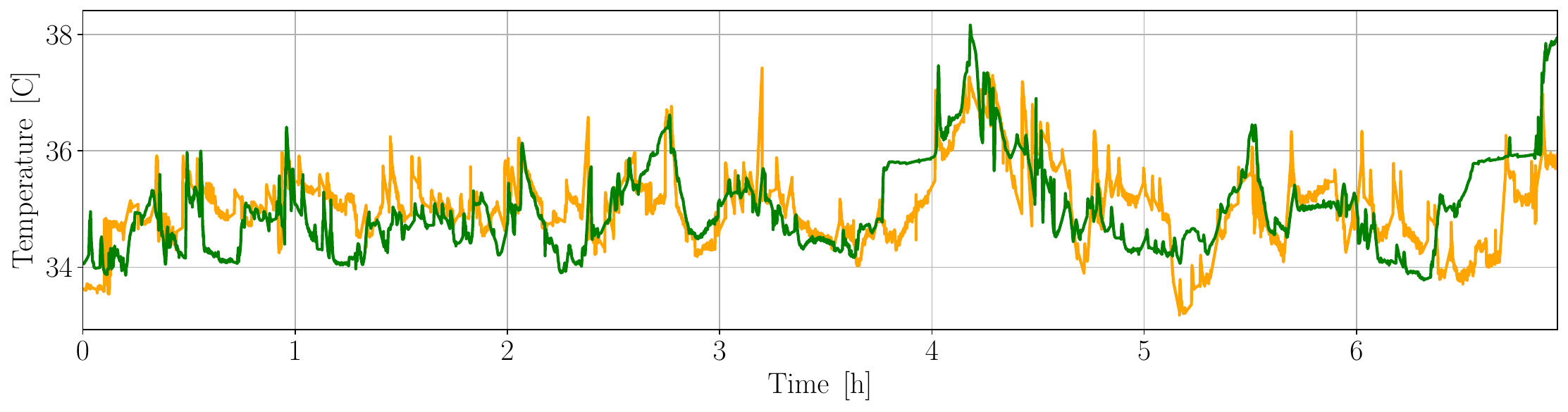}
        \\[1ex] % Add some vertical space
    \end{minipage}\\[1ex] % Add some vertical space
    \caption{Pixel predictions of the 10 optimal sensor positions (\protect\greenline) compared to the ground truth (\protect\orangeline).}
    \label{fig:LSTM_measurement_predictions}
\end{figure*}

\section{Conclusion}
In conclusion, the application of Robust Principal Component Analysis (RPCA) on thermal image data significantly enhances the quality of the data, allowing for more insightful subsequent analyses. Given its robustness and versatility, this method can be extended to various data applications, broadening its relevance and potential impact across diverse domains. Furthermore, the use of Optimal Sensor Placement (OSP) offers a promising approach for those looking to maximize the efficiency of their data storage and compression strategies, especially in environments where storage space and data transmission capabilities are limited. Applying LSTMs to a lower-dimensional space obtained by OSP can improve computational efficiency and can enhance the accuracy of time-series predictions. The interaction of the presented approaches optimizes both data processing and subsequent analyses, which can improve data quality, computational efficiency, and memory efficiency while enabling real-time predictive capabilities.

\bibliographystyle{AR}
\bibliography{references}

\end{document}